\documentclass{article}

\PassOptionsToPackage{numbers,square,sort&compress}{natbib}
\usepackage[main,final]{neurips_2026}
\usepackage{microtype}
\usepackage{amsmath,amssymb}
\usepackage{amsfonts}
\usepackage{amsthm}
\usepackage{bbm}

\usepackage{algorithm}
\usepackage{algpseudocode}
\usepackage{float}

\algnewcommand\Input{\item[\textbf{Input:}]}
\algnewcommand\Output{\item[\textbf{Output:}]}

\usepackage[table]{xcolor}
\usepackage{graphicx}
\usepackage{booktabs}
\usepackage{adjustbox}
\usepackage{wrapfig}
\usepackage{enumitem}
\usepackage{tikz-cd}
\usepackage{multicol}
\usepackage{multirow}
\usepackage{subcaption}
\usepackage{nicefrac}

\graphicspath{{Figure/}{../Figure/}}

\usepackage[utf8]{inputenc}
\usepackage[T1]{fontenc}
\usepackage{microtype}
\usepackage{url}
\usepackage{hyperref}

\usepackage{titletoc}

\usepackage{kotex}

\newcommand{\meanstdsmall}[2]{$#1\raisebox{-0.0ex}{\scriptsize $\pm #2$}$}
\newcommand{\fix}[1]{\textcolor{black}{#1}}

\newcommand{\calibPosthoc}{
guo2017calibration,
zhang2020mix,
yang2023beyond,
kull2019beyond,
hwang2025t,
zadrozny2002transforming,
zadrozny2001obtaining,
platt1999probabilistic,
choi2024consistency}

\newcommand{\calibTraintime}{
szegedy2016rethinking,
zhang2017mixup,
lin2017focal,
lakshminarayanan2017simple,
liang2020improved,
hebbalaguppe2022stitch,
pereyra2017regularizing,
chu2025unlocking}

\theoremstyle{definition}

\title{
MT-CC: Multi-Group Temperature Scaling for
Asymmetric Calibration Behavior in Class-Incremental Learning
}

\author{%
  Sungjun Yun \quad \quad
  Yonghee Choi \quad \quad
  Dong-Jun Han\\[4pt]%\textsuperscript{\dag} \\[4pt]
  Yonsei University \\[4pt]
  \texttt{\{sungjun.yun, 2025321337, djh\}@yonsei.ac.kr}
}

\begin{document}

\maketitle
\begingroup
\renewcommand{\thefootnote}{}
%\footnotetext{\textsuperscript{\dag}Corresponding author.}
\footnotetext{Code available at https://github.com/DaeZZang/MT-CC}
\endgroup
\begin{abstract}
    As deep neural networks are required to continuously adapt to evolving data, class-incremental learning (CIL) has become an active research area. However, most existing works primarily focus on improving accuracy while overlooking confidence calibration, which measures the trustworthiness of the model's predictions. Although temperature scaling (TS) is effective for static calibration, it has rarely been explored in continual settings where data distributions evolve over time. In this paper, we introduce Multi-Group Temperature Scaling for Continual Calibration (MT-CC), a novel post-hoc calibration framework that addresses the asymmetric calibration behavior in CIL, where the rate of confidence changes does not align with the rate of accuracy degradation across tasks, leading to different levels of miscalibration. Our key contribution is to design multiple temperature parameters through a lightweight grouping network that separates samples by confidence statistics and task-aware signals in CIL. Furthermore, a Groupwise Difference-between-Confidence-and-Accuracy (GDCA) regularization is incorporated to promote intra-group consistency. Experiments demonstrate that MT-CC significantly improves calibration performance while maintaining accuracy, offering a principled framework for reliable continual learning.
\end{abstract}

\section{Introduction}
\label{sec:intro}

In modern deep learning applications, it has become increasingly important for a model to provide reliable indications of whether its predictions are likely to be correct. In particular, in a well-calibrated model, the predicted confidence should align with the empirical likelihood that the prediction is correct. This allows users to trust the model’s outputs when it exhibits high confidence, while rejecting predictions made with low confidence. 
However, in practice, neural networks are often miscalibrated, producing confidence scores that do not align with their actual accuracy, frequently being overconfident in wrong predictions. 

In this context, confidence calibration has attracted increasing attention, aiming to align the model's predicted confidence with the true likelihood of correctness. Various methods have been proposed, including temperature scaling (TS)~\cite{guo2017calibration}, which performs a post-hoc adjustment of logits to improve calibration without altering the model’s predictions. In addition to such post-hoc approaches \cite{\calibPosthoc}, train-time calibration methods have also been explored \cite{\calibTraintime}, where the model is explicitly trained to produce well-calibrated confidence scores. Compared to train-time schemes, post-hoc methods such as TS offer a clear advantage, as they can be directly applied to any pre-trained model without sacrificing accuracy.

\begin{figure}
    \centering
    \includegraphics[width=1.0\linewidth]{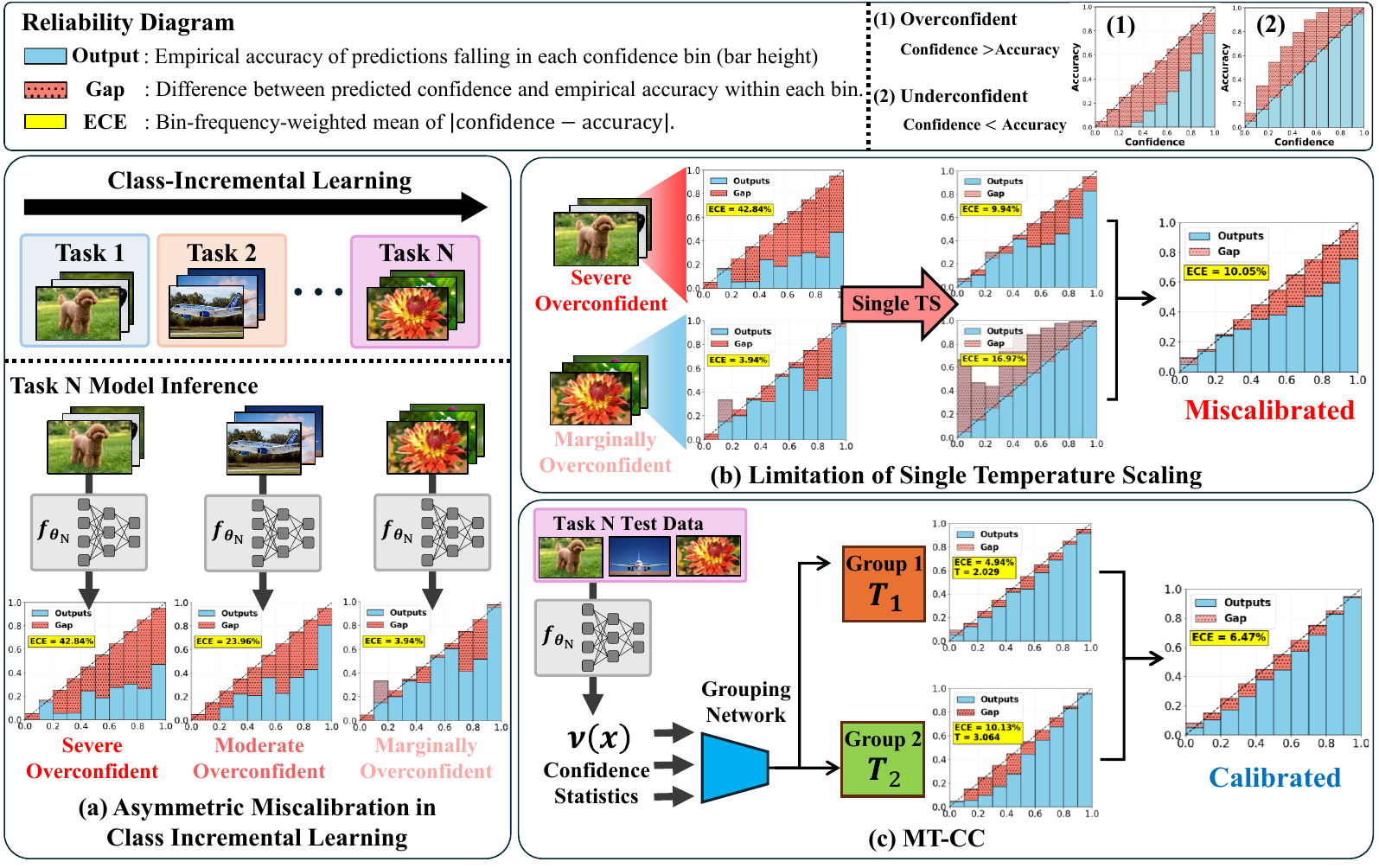}
    \caption{\small The reliability diagram visualizes calibration by comparing empirical accuracy and predicted confidence within each confidence bin, while a lower ECE indicates better calibration. \textbf{(a)} In class-incremental learning, accuracy and confidence evolve at \emph{different rates} across tasks, producing \textbf{asymmetric calibration behaviors}: the gap between confidence and accuracy varies across tasks due to catastrophic forgetting.
  \textbf{(b)} As a result, a single global temperature cannot simultaneously calibrate these heterogeneous groups.
 \textbf{(c)} Our MT-CC routes each sample to a group-specific temperature to capture asymmetric calibration behaviors in class-incremental learning.}
    \label{fig:overview}
    \vspace{-7mm}
\end{figure}

\noindent \textbf{Fundamental challenges.} However, most prior works on calibration have focused on static scenarios, where the target task and data distribution remain fixed. In many practical settings, however, models are required to learn continuously from a stream of tasks where new classes are introduced over time without revisiting all previous data. In such dynamic environments, maintaining well-calibrated confidence becomes particularly challenging, as the distribution shifts induced by new tasks often lead to varying levels of miscalibration across tasks.

\begin{wrapfigure}[15]{r}{0.43\textwidth}
\centering
\vspace{-5mm}
\includegraphics[width=\linewidth]{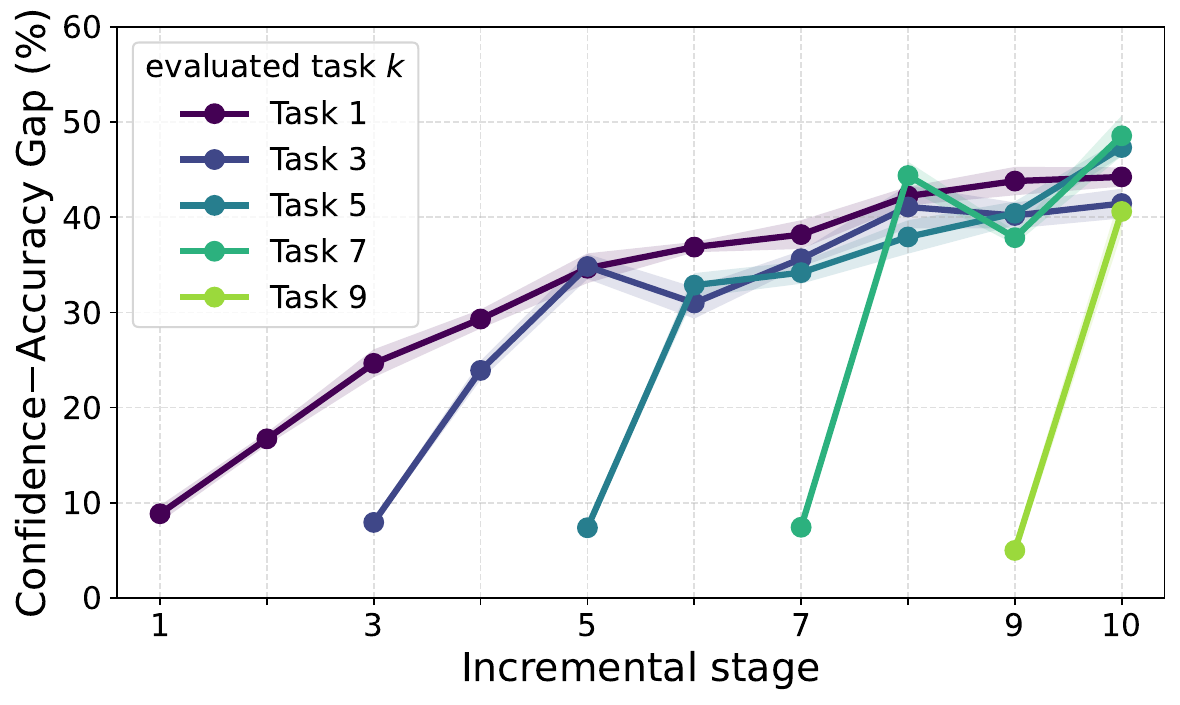}
\vspace{-6mm}
\caption{\small
Confidence-accuracy gap of selected tasks over incremental stages of CIFAR-100. 
The results reveal \textbf{asymmetric calibration behaviors} in continual learning, where the discrepancy between confidence and accuracy varies across tasks due to catastrophic forgetting. Results for all tasks are provided in
Appendix~\ref{sec:forgetting_asymmetric_calibration}.
}
\label{fig:last_task_acc_conf_cifar100}
\vspace{-4mm}
\end{wrapfigure}

Concretely, as illustrated in Figure~\ref{fig:overview}a,
models in class-incremental learning (CIL)~\citep{wu2019large,parisi2019continual,de2021continual}
exhibit asymmetric calibration behaviors, where accuracy and confidence evolve at different rates across tasks under continual updates (see Figure~\ref{fig:last_task_acc_conf_cifar100}).
As a result, a single temperature value $T$ fails to calibrate all classes effectively, since different task groups exhibit distinct confidence characteristics and require differentiated scaling.

Although T-CIL \citep{hwang2025t} recently proposed a TS strategy tailored to CIL, it still relies on a single temperature value and thus cannot effectively address the asymmetric calibration behaviors of CIL, as illustrated in Figure~\ref{fig:overview}b. 
\citet{tomani2022parameterized,joy2023sample} propose sample-dependent temperature scaling, but they are developed under static, fixed-distribution settings and do not explicitly capture the CIL-specific calibration heterogeneity induced by continual task shifts. To the best of our knowledge, these aspects have been largely overlooked in existing calibration research and remain unresolved. Our work seeks to bridge that gap by addressing the following question:

\noindent \textit{How can we ensure reliable model calibration in continual learning, where asymmetric miscalibration behaviors across tasks pose persistent challenges?}

\noindent \textbf{Contributions.} In this paper, we propose multi-group temperature scaling for continual calibration (MT-CC), a novel calibration strategy that introduces multiple temperature parameters determined by a grouping network, effectively tackling the asymmetric calibration behavior in continual learning. 
To this end, we train the grouping network using CIL-specific confidence statistics, including the probability mass assigned to old and new classes and the logit margin, which together capture heterogeneous calibration needs.  
This enables the grouping network to assign an appropriate temperature to each group, thereby achieving calibration both within each group and across the entire dataset, as shown in Figure~\ref{fig:overview}c.
In addition, we propose a Groupwise Difference-between-Confidence-and-Accuracy (GDCA) loss to balance confidence and accuracy within each group, thereby further improving calibration.

\noindent Our main contributions are summarized as follows:

\begin{itemize}
[topsep=0pt,itemsep=4pt,parsep=0pt, partopsep=0pt, leftmargin=*]
 \item 
     We analyze asymmetric calibration behaviors in class-incremental learning, where different tasks exhibit varying degrees of miscalibration, and exploit these behaviors to design a CIL-aware calibration framework.
    \item 
     We propose multi-group temperature scaling for continual calibration (MT-CC), a novel calibration framework that introduces multiple temperature parameters determined by a grouping network, enabling input-dependent calibration. The grouping network is conditioned on old/new confidence statistics together with a logit margin signal, allowing it to capture asymmetric calibration behaviors and assign an appropriate temperature to each group.
    \item 
    We conduct extensive experiments using various benchmark datasets, demonstrating the superiority of MT-CC in class-incremental settings.
\end{itemize}

\vspace{-2mm}
\section{Related Work}\label{sec:related}
\vspace{-2mm}

\noindent \textbf{Confidence calibration.} Compared to train-time calibration methods \citep{\calibTraintime} that incorporate calibration objectives during the training process, post-hoc methods \citep{\calibPosthoc} calibrate the confidence by adjusting the output logits of an already trained model. Post-hoc methods offer a clear advantage in that they preserve accuracy and can be directly applied to pre-trained models.
A representative approach is temperature scaling (TS), which rescales logits with a single scalar temperature~\citep{guo2017calibration}.
Subsequent extensions, including ETS, IRM~\citep{zhang2020mix}, PTS~\citep{tomani2022parameterized}, AdaTS~\citep{joy2023sample}, and grouping calibration~\citep{yang2023beyond}, relax the single-temperature assumption of TS.
However, they are primarily designed for static settings and thus struggle to handle continual task shifts in CIL.
As shown in Section~\ref{sec:exp}, directly applying these methods to CIL yields limited performance because they overlook the asymmetric calibration behavior induced by continual updates and catastrophic forgetting. MT-CC instead learns a grouping function from CIL-aware signals, including old/new confidence and logit margin: while the former directly captures the old/new class structure unique to CIL, the latter becomes CIL-relevant because catastrophic forgetting alters its relationship with miscalibration relative to static settings.

\noindent \textbf{Class-incremental learning.}
The central challenge in CIL is catastrophic forgetting, where learning new classes degrades performance on previously learned ones~\citep{wu2019large,parisi2019continual,de2021continual,masana2022class}. 
To address this, prior work has proposed regularization, parameter-isolation, and replay-based methods~\citep{kirkpatrick2017overcoming,zenke2017continual,aljundi2018memory,zhao2020maintaining,yan2021dynamically,castro2018end,li2017learning,kim2023warping,rusu2016progressive,serra2018overcoming,mallya2018piggyback,chaudhry2019tiny,lopez2017gradient,rebuffi2017icarl,chaudhry2018efficient,aljundi2019online}, all aiming to preserve past knowledge and maintain accuracy across learned classes.
However, most CIL studies focus primarily on classification accuracy, while largely overlooking confidence calibration.

\noindent \textbf{Confidence calibration in continual learning.}
\citet{li2024calibration} recently studied confidence calibration in continual learning.
T-CIL~\citep{hwang2025t}, a recent state-of-the-art calibration method tailored to CIL, estimates a temperature after each task using augmented replay data from validation samples.
However, its single-temperature formulation struggles to capture the asymmetric calibration behaviors that emerge across tasks under continual updates.
%\djh{However, its single-temperature formulation struggles to capture the asymmetric calibration behaviors that emerge across samples and tasks under continual updates.}
MT-CC addresses this limitation by assigning multiple temperatures through a grouping network for task-aware calibration.
\vspace{-1mm}

\section{Background and Problem Setup}\label{sec:background}
\vspace{-1mm}

\subsection{Background: Confidence Calibration and Temperature Scaling}

\vspace{-1mm}

\noindent\textbf{\fix{Confidence calibration.}} In a $K$-way classification problem with label space $\mathcal{Y} = \{1, \dots, K\}$, a model parameterized by $\theta$ outputs a probability vector $s = \mathrm{softmax}(f_\theta(x)) \in [0,1]^K$ for an input $x$, where $f_\theta(x) \in \mathbb{R}^K$ denotes the logit vector. The predicted label is $\hat{y} = \arg\max_{y \in \mathcal{Y}} s[y]$, where $s[y]$ denotes the predicted probability for class $y \in \mathcal{Y}$, and its associated confidence is $\hat{s} = \max_{y \in \mathcal{Y}} s[y]$. A model is perfectly calibrated if $\mathbb{P}(y = \hat{y} \mid \hat{s} = p) = p$ for all $p \in [0,1]$, meaning that predictions with confidence $p$ are correct with probability $p$.

\noindent\textbf{\fix{Temperature scaling.}}
Temperature scaling (TS)~\citep{guo2017calibration} is a post-hoc method that achieves calibration by rescaling the model's output logit ${z}$ using a single scalar temperature $T>0$.
Given a sample $x$ and its corresponding logit $z = f_{\theta}(x) \in \mathbb{R}^K$, 
the calibrated probability vector for sample $x$ is computed as $s(z,T)
= \mathrm{softmax}\!\left(\frac{z}{T}\right)$.
The temperature $T$ is learned on a validation set $\mathcal{D}^{\mathrm{val}} = \{({x}_i, y_i)\}_{i=1}^{N_{\text{val}}}$ by minimizing the negative log-likelihood (NLL) loss.

\vspace{-1mm}
\subsection{Problem Setup: Calibration in Class-Incremental Learning}
\vspace{-1mm}
\textbf{Class-incremental learning setup.} We consider the standard class-incremental learning (CIL) setting, consisting of a sequence of $B$ tasks  
$\{\mathcal{D}^1, \mathcal{D}^2, \dots, \mathcal{D}^B\}$,  
where each task $\mathcal{D}^b = \{({x}_i^b, y_i^b)\}_{i=1}^{n_b}$  
contains $n_b$ training samples.  
Here, $x_i^b$ denotes an input sample  
and $y_i^b \in \mathcal{Y}_b$ represents its corresponding label. 
Each task has a distinct label space,  
i.e., $\mathcal{Y}_b \cap \mathcal{Y}_{b'} = \emptyset$ for $b \neq b'$.  After completing task $b$,  
the model is evaluated on all classes encountered so far,  
denoted as $\mathcal{Y}_{1:b} = \mathcal{Y}_1 \cup \mathcal{Y}_2 \cup \dots \cup \mathcal{Y}_b$.  
Following \citet{hwang2025t},  
we maintain a small memory buffer $\mathcal{M}_{b-1}$  
storing representative samples from previous tasks to mitigate catastrophic forgetting.  
During training on task $b$,  
the model learns jointly from the current task data $\mathcal{D}^b$ and the memory buffer $\mathcal{M}_{b-1}$.  
After training, the buffer is updated to $\mathcal{M}_b$ by retaining a subset of samples from $\mathcal{D}^b \cup \mathcal{M}_{b-1}$.

\noindent \textbf{Goal and problem formulation.} Unlike prior works that primarily focus on improving accuracy, we aim to address confidence calibration in CIL explicitly. To this end, we integrate TS into the CIL framework by optimizing the temperature after each task $b$ and applying it to calibrate model confidence. Specifically, each task $b$ consists of two steps:

\vspace{-2mm}
\begin{itemize}[leftmargin=*, labelsep=0.5em]
    \item \textbf{Training step.}  
    The model $\theta$ learns on $\mathcal{D}^b \cup \mathcal{M}_{b-1}$ 
    to acquire new knowledge while preserving performance on previous tasks.

    \item \textbf{Calibration step.}  
   After training is completed, the model is recalibrated using validation data $\mathcal{D}_{b}^{\text{val}}$, adjusting its confidence mapping to reflect the new data distribution.
\end{itemize}

\vspace{-2mm}
Our goal is to build a well-calibrated model that remains reliable as tasks evolve, thereby reducing miscalibration arising from distributional and label-space shifts.

\vspace{-1mm}
\subsection{Aren’t the Existing Solutions Good Enough?}
\vspace{-1mm}

In CIL, we identify a key challenge that prior works in static settings have overlooked: the confidence-accuracy gap generally widens under continual updates,
with different rates and trajectories across tasks, 
resulting in \textit{different levels of miscalibration} (Figure~\ref{fig:last_task_acc_conf_cifar100}). Consequently, conventional TS methods that employ a single global temperature are unable to simultaneously achieve good calibration across heterogeneous confidence distributions.
As illustrated in Figure~\ref{fig:overview}b, the single-temperature approach in T-CIL~\cite{hwang2025t}, the current state-of-the-art calibration method for CIL, fails to maintain consistent calibration, highlighting its limitation in handling asymmetric miscalibration behaviors. Moreover, while several adaptive per-sample temperature scaling methods have been proposed \cite{tomani2022parameterized,joy2023sample}, they typically assume a static data distribution and do not explicitly account for continual task evolution. We later show that these prior works exhibit significant miscalibration in CIL settings.
These observations underscore the need for a more flexible, group-aware calibration strategy tailored to CIL.

\vspace{-1mm}
\section{Proposed MT-CC Algorithm}\label{sec:algo}
\vspace{-1mm}

We propose MT-CC (\textit{Multi-Group Temperature Scaling for Continual Calibration}), 
a new calibration framework designed for class-incremental learning.

\noindent \textbf{Overview of approach.} 
At its core, MT-CC introduces a lightweight grouping module (Sections \ref{sec:grouping} and \ref{sec:training_group}) that learns to partition samples into multiple groups 
based on confidence- and margin-based indicators, reflecting heterogeneous miscalibration behaviors in CIL. 
Each group is then equipped with its own temperature parameter, allowing calibration to adapt locally rather than relying on a single global scaling factor. 
In addition, a Groupwise Difference-between-Confidence-and-Accuracy (GDCA) loss (Section \ref{sec:gdca}) is employed to align predicted confidence with correctness within each group, further ensuring intra-group reliability and enhancing overall calibration. 

\vspace{-1mm}

\begin{figure*}[t]
    \centering
    \begin{subfigure}[t]{0.40\textwidth}
        \centering
        \includegraphics[
            width=\linewidth,
            height=0.24\textheight,
            keepaspectratio
        ]{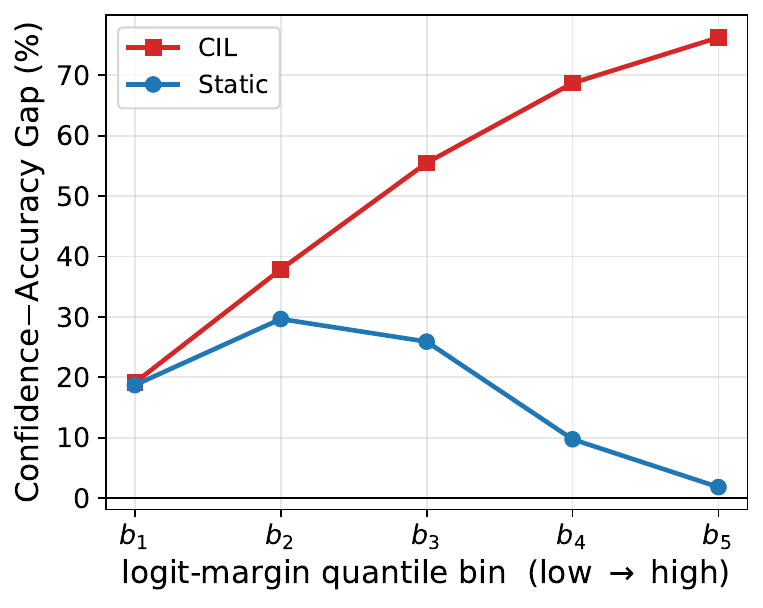}
        \caption{
        Confidence--accuracy gap across logit-margin quantile bins on CIFAR-100.
        }
        \label{fig:margin_calibration}
    \end{subfigure}
    \hfill
    \begin{subfigure}[t]{0.56\textwidth}
        \centering
        \includegraphics[
            width=\linewidth,
            height=0.24\textheight,
            keepaspectratio
        ]{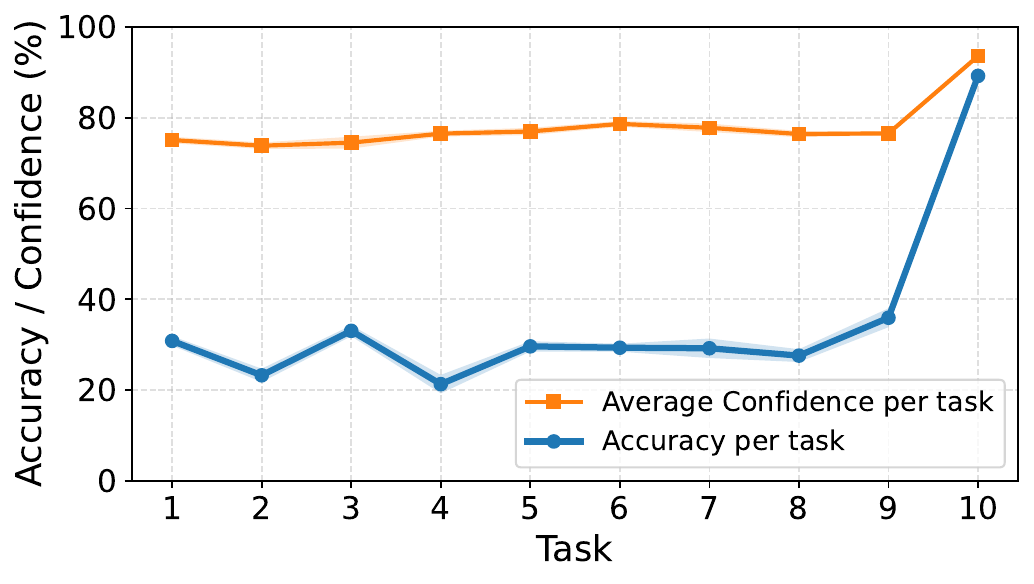}
        \caption{
        Task-wise accuracy and confidence on CIFAR-100 after learning all 10 tasks.
        }
        \label{fig:last_task_acc_conf_cifar100_2}
    \end{subfigure}
    \caption{\small
    \textbf{Calibration behavior in CIL.}
    (a) The final-stage CIL model is compared with a static model jointly trained on all classes using five equal-frequency logit-margin bins from low to high margin. The confidence--accuracy gap increases with margin only under CIL.
    (b) Task-wise accuracy and confidence are measured after learning all 10 tasks, showing larger gaps for earlier tasks than for newly introduced tasks.
    }
    \label{fig:cil_calibration_behavior}
    \vspace{-3mm}
\end{figure*}
\subsection{Grouping Network Design}
\label{sec:grouping}

We define a set of group-specific temperature parameters $\{T_1, T_2, \dots, T_M\}$, where $M$ denotes the total number of groups. We then introduce a grouping network $g_{\phi}$, whose goal is to map each sample 
to the most appropriate temperature according to its confidence-related characteristics.  

\noindent{\textbf{Grouping network mechanism (Inference).}} The grouping network $g_{\phi}$ is applied after the classifier, and takes the model’s logits along with derived confidence statistics as inputs.  
For each sample ${x}$ with classifier output ${z} = f_{\theta}({x}) \in \mathbb{R}^{K}$,  
we construct a calibration feature vector $v({x})$ as: 
\begin{equation}
v({x}) = [{\delta},\, p_{\text{old}},\, p_{\text{new}},\, \,p\,],
\label{eq:calibration_feature_vector}
\end{equation}
where $v({x})$ concatenates informative signals capturing different aspects of model confidence and uncertainty.  
Specifically, $v(x)$ consists of the softmax probability vector $p=\mathrm{softmax} (z)$, the logit margin ${\delta}$, the old-task confidence $p_{\text{old}}$, and the new-task confidence $p_{\text{new}}$.
Then, the concatenated vector $v(x)$ passes through the grouping network $g_{\phi}$ as follows: 
$g_{\phi}(x) = \mathrm{softmax}(W_{\phi} v(x) + b_{\phi})$,
where $W_{\phi} \in \mathbb{R}^{M \times d}$ and $b_{\phi} \in \mathbb{R}^{M}$  
are learnable parameters of the grouping network, $M$ is the total number of calibration groups,  
and $v(x) \in \mathbb{R}^{d}$ denotes the calibration feature vector in Eq.~\ref{eq:calibration_feature_vector}. The output of the grouping function, i.e., $g_{\phi}(x)$, produces a probability distribution over the $M$ calibration groups. Consequently, the sample $x$ is assigned to the group with the highest probability (e.g., group $m$), and the final calibrated output is given by $p_{\mathrm{cal}}(x)=\mathrm{softmax}\!\left(\frac{f_{\theta}(x)}{T_{m}}\right)$. Here, $T_m$ denotes the temperature associated with the group to which sample $x$ is assigned by $g_{\phi}(x)$.

\noindent{\textbf{Design of grouping network inputs.}}
One of our key contributions is the CIL-aware input design in Eq.~\ref{eq:calibration_feature_vector}. The input features are designed to capture complementary aspects of prediction confidence under continual learning: the logit margin tracks overconfidence caused by forgetting under CIL, the old/new-task confidences explicitly summarize the old-versus-new class structure that is otherwise implicit in the full softmax vector, while the softmax vector preserves class-wise detail. Together, these signals allow the grouping network to distinguish samples with different calibration needs.

\noindent\textbullet\hspace{0.3em}\textbf{Logit margin ($\delta$):}
We define the logit margin as the difference between the two largest logits,
$\delta(x)=z_{(1)}-z_{(2)}$.
To analyze calibration behavior across logit margins, we compare the final-stage CIL model with a static model jointly trained on all CIFAR-100 classes. Using the uncalibrated outputs of each model on the same test set, we divide samples into five equal-frequency logit-margin bins, $b_1,\ldots,b_5$,  
ordered from low to high margin. We report the confidence–accuracy gap in each bin, averaged over five seeds.
As shown in Figure~\ref{fig:margin_calibration}, unlike in the static setting, overconfidence under CIL increases systematically with the logit margin.
Consistently, the NLL-optimal temperature increases with $\delta$ under CIL, while remaining nearly constant in the static model (Appendix~\ref{sec:margin_temperature}), indicating that high-margin predictions require stronger calibration correction under CIL. 
We exploit this relationship to predict calibration needs, using $\delta$ as a CIL-aware input signal. 

\noindent\textbullet\hspace{0.3em}\textbf{Old/new-task confidence ($p_{\text{old}}, p_{\text{new}}$):} For a given sample $x$, we define the old-task confidence as the total probability mass assigned to previously learned classes, $p_{\text{old}}(x) = \sum_{y \in \mathcal{C}_{\text{old}}} p(y \mid x)$, where $p(y \mid x)$ denotes the softmax probability for class $y$. If the current task is $b$, the old-class set is given by
$\mathcal{C}_{\text{old}} = \mathcal{Y}_{1:b-1}$.
Similarly, we define the new-task confidence as $p_{\text{new}}(x) = \sum_{y \in \mathcal{C}_{\text{new}}} p(y \mid x)$, where $\mathcal{C}_{\text{new}} = \mathcal{Y}_{b}$ denotes the set of classes introduced in the current task. 
As shown in Figure~\ref{fig:last_task_acc_conf_cifar100_2}, after all 10 tasks, old tasks generally exhibit larger confidence--accuracy gaps than new tasks, revealing distinct calibration behavior.
We therefore use their respective probability masses as CIL-aware signals to capture sample-specific calibration needs.
These quantities explicitly summarize the old-versus-new class structure contained in the full softmax vector, directly exposing this CIL-relevant structure to the grouping network. 
% As shown in Appendix~\ref{sec:explicit_old_new}, this explicit structure consistently improves calibration over using only \([\delta,p]\) across different routing-training data sizes, while comparable random class-mass summaries do not provide the same benefit.
Appendix~\ref{sec:explicit_old_new} shows consistent improvements over using only \([\delta,p]\) across routing-training data sizes, whereas random class-mass summaries yield no comparable benefit.

\noindent\textbullet\hspace{0.3em}\textbf{Softmax probability vector ($p$):}
We use the full softmax probability vector,
$p=\mathrm{softmax}(f_{\theta}(x))$, to capture fine-grained class-wise confidence information. Even when samples have similar logit margins, their probability distributions across classes can differ, providing additional information for identifying different calibration needs. We further analyze this complementarity in Appendix~\ref{sec:margin_confidence_complementarity}.  

\subsection{Training Grouping Network}
\vspace{-1mm}
\label{sec:training_group}
At the beginning of each task, the validation set is divided into two disjoint subsets of equal size: 
 $\mathcal{S}_{\text{grp}}$ is used to train the grouping network $g_{\phi}$,  
while $\mathcal{S}_{\text{cal}}$ is the set reserved for groupwise temperature scaling in the subsequent stage.
This split prevents overfitting and ensures that temperature parameters are tuned on unseen calibration samples.  

\noindent \textbf{Grouping network training.} During optimization, $g_{\phi}$ and the group temperatures $\{T_1, T_2, \dots, T_M\}$  
are trained jointly on $\mathcal{S}_{\text{grp}}$ under the following objective: 
\begin{equation}
\mathcal{L}_{\text{MT-CC}}
=
-\frac{1}{|\mathcal{S}_{\text{grp}}|}
\sum_{(x,y)\in \mathcal{S}_{\text{grp}}}\log
\sum_{m=1}^{M}
g_{\phi}(x)_m
\cdot
s(f_{\theta}(x),T_m)[y].
\label{eq:mtcc}
\end{equation}
This objective is designed as a differentiable relaxation of hard groupwise temperature scaling. Ideally, each sample should be assigned to a single calibration group, but directly optimizing such discrete assignments is non-differentiable. We therefore use the soft routing weights $g_{\phi}(x)_m$ during this stage so that the grouping network and temperatures can be optimized jointly.

\noindent \textbf{Groupwise temperature scaling.}
After training, the grouping network $g_{\phi}$ partitions $\mathcal{S}_{\text{cal}}$ into $M$ disjoint groups according to the most probable assignment.  
Each sample $x$ is thereby assigned to a single group  
and temperature scaling is applied independently within each group. Following the standard temperature scaling, the optimal temperature $T_m^{*}$ for group $m$ is obtained by minimizing the NLL loss over its corresponding subset. Each group’s initial temperature $T_m$ is initialized with the value learned during the joint MT-CC optimization, which provides a well-informed starting point for fine-tuning on its local calibration subset.  
The resulting groupwise temperatures $\{T_m^{*}\}$ are then stored and reused as initialization in the next task’s MT-CC stage. 

\vspace{-2mm}
\subsection{Groupwise DCA Loss}\label{sec:gdca}
\vspace{-1mm}

We introduce the \emph{Groupwise Difference-between-Confidence-and-Accuracy (GDCA)} loss as an auxiliary objective.
The standard DCA loss penalizes the gap between the average confidence and empirical accuracy of a model~\citep{liang2020improved}.
We extend this idea to the group level by enforcing such alignment within each calibration group.
Let $G_m$ denote the subset of samples assigned to the $m$-th calibration group.
The GDCA loss is defined as
\begin{equation}
\mathcal{L}_{\text{GDCA}}
=
\sum_{m=1}^{M}
w_m
\left|
\operatorname{Acc}(G_m) - \operatorname{Conf}(G_m)
\right|,
\qquad
w_m =
\frac{\sqrt{|G_m|}}{\sum_{j=1}^{M}\sqrt{|G_j|}},
\label{eq:gdca}
\end{equation}
where $\operatorname{Acc}(G_m)$ and $\operatorname{Conf}(G_m)$ denote the empirical accuracy and average calibrated confidence within group $G_m$, respectively.
Here, $\operatorname{Conf}(G_m)$ is computed using the calibrated mixture confidence
$a(x,\hat{y}) \triangleq \sum_{k=1}^{M} g_{\phi}(x)_k\, s(z,T_k)[\hat{y}]$ for the predicted class $\hat{y}$.
GDCA gradients are carried by the confidence term, updating the temperature parameters. Through these shared temperatures, GDCA stabilizes the coupled routing-temperature optimization and promotes groupwise calibration consistency (see Appendix~\ref{sec:gdca_stabilization} for a detailed mechanism-level discussion).
The overall training objective of MT-CC is written as
$\mathcal{L}_{\text{total}} = \mathcal{L}_{\text{MT-CC}} + \lambda \mathcal{L}_{\text{GDCA}}$, where $\lambda$ controls the strength of GDCA.

\vspace{-2mm}
\section{Experimental Results}\label{sec:exp}
\vspace{-2mm}

We evaluate the proposed MT-CC framework by measuring the calibration performance under the CIL setting. All results are reported over five seeds for CIFAR-10, CIFAR-100, Tiny-ImageNet, and ViT-based CIL experiments, and over three seeds for ImageNet.

\noindent\textbf{Datasets and CIL setup.}
We conduct experiments on the widely used benchmarks for CIL: CIFAR-10 \citep{krizhevsky2009learning}, CIFAR-100 \citep{krizhevsky2009learning}, Tiny-ImageNet \citep{le2015tiny} and ImageNet \citep{deng2009imagenet}.  
Following the training protocol of T-CIL~\citep{hwang2025t}, for CIFAR-10 and CIFAR-100, we use a ResNet-32 backbone \citep{he2016deep} with 5 tasks (2 classes per task) and 10 tasks (10 classes per task), retaining memory sizes of 200 and 2{,}000 samples, respectively. For Tiny-ImageNet, we adopt ResNet-18 and construct 10 tasks of 20 classes each, using 2{,}000 memory samples. \fix{For ImageNet, we employ ResNet-18 and partition the 1{,}000 classes into 10 tasks, forming a challenging setup with memory sizes of 5{,}000, 10{,}000, and 20{,}000 samples.}
We also evaluate ViT-B/16 \citep{dosovitskiy2020image} with ER on CIFAR-100 and ImageNet-R \citep{hendrycks2021many}, using 10 tasks, a 2{,}000-sample replay buffer, and 500 validation samples.

\begin{table*}[t]
% \vspace{-4mm}
\centering
\caption{\fix{Comparison with various baseline methods. Since all evaluated methods are post-hoc calibration techniques, they preserve the original classification accuracy of the base models (65.24\% for CIFAR-10, 56.47\% for CIFAR-100, and 29.42\% for Tiny-ImageNet). Each performance value is averaged over all sequential tasks in CIL. While maintaining identical classification performance, our method robustly outperforms the SOTA T-CIL in every calibration metric across diverse setups.}}
\vspace{-2mm}
\label{tab:main_results}
\setlength{\tabcolsep}{2pt}
\begin{adjustbox}{max width=0.85\textwidth}
\begin{tabular}{@{}l|ccc|ccc|ccc@{}}
\toprule
\textbf{Dataset}
& \multicolumn{3}{c|}{\textbf{CIFAR-10}}
& \multicolumn{3}{c|}{\textbf{CIFAR-100}}
& \multicolumn{3}{c}{\textbf{Tiny-ImageNet}} \\
\cmidrule(lr){2-10}
\textbf{Method}
& \textbf{ECE ($\downarrow$)} & \textbf{AECE ($\downarrow$)} & \textbf{SCE ($\downarrow$)}
& \textbf{ECE ($\downarrow$)} & \textbf{AECE ($\downarrow$)} & \textbf{SCE ($\downarrow$)}
& \textbf{ECE ($\downarrow$)} & \textbf{AECE ($\downarrow$)} & \textbf{SCE ($\downarrow$)} \\
\midrule
\midrule
Vanilla      
& \meanstdsmall{28.51}{0.85} & \meanstdsmall{28.49}{0.86} & \meanstdsmall{8.83}{0.31}
& \meanstdsmall{27.59}{0.11} & \meanstdsmall{27.57}{0.11} & \meanstdsmall{1.35}{0.02}
& \meanstdsmall{45.72}{0.16} & \meanstdsmall{45.72}{0.16} & \meanstdsmall{1.28}{0.01} \\
TS \citep{guo2017calibration}           
& \meanstdsmall{24.27}{0.84} & \meanstdsmall{24.24}{0.85} & \meanstdsmall{8.16}{0.30}
& \meanstdsmall{16.13}{0.19} & \meanstdsmall{16.09}{0.23} & \meanstdsmall{0.99}{0.01}
& \meanstdsmall{27.28}{0.58} & \meanstdsmall{27.30}{0.60} & \meanstdsmall{0.93}{0.01} \\
ETS \citep{zhang2020mix}          
& \meanstdsmall{22.75}{0.83} & \meanstdsmall{22.65}{0.81} & \meanstdsmall{5.92}{0.20}
& \meanstdsmall{16.70}{0.39} & \meanstdsmall{17.62}{0.25} & \meanstdsmall{0.56}{0.01}
& \meanstdsmall{28.93}{0.59} & \meanstdsmall{28.90}{0.63} & \meanstdsmall{0.52}{0.01} \\
IRM \citep{zhang2020mix}          
& \meanstdsmall{22.16}{0.69} & \meanstdsmall{22.01}{0.65} & \meanstdsmall{5.85}{0.16}
& \meanstdsmall{17.64}{0.37} & \meanstdsmall{17.51}{0.39} & \meanstdsmall{0.56}{0.01}
& \meanstdsmall{28.35}{0.84} & \meanstdsmall{28.32}{0.80} & \meanstdsmall{0.52}{0.01} \\
PTS \citep{tomani2022parameterized}          
& \meanstdsmall{25.24}{2.01} & \meanstdsmall{25.10}{1.99} & \meanstdsmall{6.08}{0.42}
& \meanstdsmall{18.48}{0.53} & \meanstdsmall{18.37}{0.53} & \meanstdsmall{0.57}{0.01}
& \meanstdsmall{36.39}{1.08} & \meanstdsmall{36.32}{1.08} & \meanstdsmall{0.56}{0.01} \\
AdaTS \citep{joy2023sample}    
& \meanstdsmall{22.45}{0.96} & \meanstdsmall{22.33}{0.97} & \meanstdsmall{5.93}{0.20}
& \meanstdsmall{17.54}{0.51} & \meanstdsmall{17.48}{0.53} & \meanstdsmall{0.56}{0.01}
& \meanstdsmall{28.77}{0.66} & \meanstdsmall{28.74}{0.70} & \meanstdsmall{0.53}{0.00} \\
\rowcolor{blue!15}\textbf{MT-CC w/o aug}
& \boldmath\meanstdsmall{21.25}{1.42} & \boldmath\meanstdsmall{21.22}{1.39} & \boldmath\meanstdsmall{5.52}{0.23}
& \boldmath\meanstdsmall{15.96}{1.05} & \boldmath\meanstdsmall{15.88}{1.09} & \boldmath\meanstdsmall{0.54}{0.01}
& \boldmath\meanstdsmall{26.00}{0.92} & \boldmath\meanstdsmall{25.97}{0.90}& \boldmath\meanstdsmall{0.51}{0.01} \\
\midrule
T-CIL \citep{hwang2025t}        
& \meanstdsmall{15.34}{1.34} & \meanstdsmall{15.32}{1.32} & \meanstdsmall{7.14}{0.38}
& \meanstdsmall{6.03}{0.42}  & \meanstdsmall{6.04}{0.35}  & \meanstdsmall{0.94}{0.03}
& \meanstdsmall{18.96}{0.35} & \meanstdsmall{18.99}{0.40} & \meanstdsmall{0.82}{0.01} \\
T-CIL + ETS  
& \meanstdsmall{13.94}{1.35} & \meanstdsmall{14.04}{1.30} & \meanstdsmall{5.38}{0.36}
& \meanstdsmall{5.63}{0.52}  & \meanstdsmall{5.84}{0.46}  & \meanstdsmall{0.53}{0.01}
& \meanstdsmall{22.14}{0.22} & \meanstdsmall{22.20}{0.22} & \meanstdsmall{0.51}{0.01} \\
T-CIL + IRM  
& \meanstdsmall{14.69}{1.16} & \meanstdsmall{14.94}{1.01} & \meanstdsmall{5.37}{0.37}
& \meanstdsmall{6.78}{0.60}  & \meanstdsmall{6.81}{0.61}  & \boldmath\meanstdsmall{0.51}{0.01}
& \meanstdsmall{17.87}{0.50} & \meanstdsmall{18.01}{0.36} & \boldmath\meanstdsmall{0.49}{0.01} \\
T-CIL + PTS          
& \meanstdsmall{15.00}{1.67} & \meanstdsmall{14.93}{1.55} & \meanstdsmall{5.32}{0.38}
& \meanstdsmall{9.87}{0.77} & \meanstdsmall{10.14}{0.67} & \meanstdsmall{0.55}{0.01}
& \meanstdsmall{21.81}{0.47} & \meanstdsmall{22.04}{0.50} & \meanstdsmall{0.51}{0.01} \\
T-CIL + AdaTS    
& \meanstdsmall{14.03}{1.59} & \meanstdsmall{14.00}{1.55} & \meanstdsmall{5.41}{0.37}
& \meanstdsmall{7.67}{0.52} & \meanstdsmall{7.66}{0.49} & \meanstdsmall{0.53}{0.01}
& \meanstdsmall{20.25}{0.58} & \meanstdsmall{20.23}{0.59} & \meanstdsmall{0.58}{0.01} \\
% \midrule
\rowcolor{blue!15}
\textbf{MT-CC}
& \boldmath\meanstdsmall{10.18}{1.48} & \boldmath\meanstdsmall{10.16}{1.54} & \boldmath\meanstdsmall{4.64}{0.25}
& \boldmath\meanstdsmall{4.49}{0.53}  & \boldmath\meanstdsmall{4.44}{0.49}  & \boldmath\meanstdsmall{0.51}{0.01}
& \boldmath\meanstdsmall{15.82}{0.29} & \boldmath\meanstdsmall{15.80}{0.32} & \meanstdsmall{0.52}{0.01} \\
\bottomrule
\end{tabular}
\end{adjustbox}
\vspace{-2mm}
\end{table*}

\begin{figure}[t]
    \centering
    \begin{subfigure}[H]{0.46\linewidth}
        \centering
        \includegraphics[width=\linewidth]{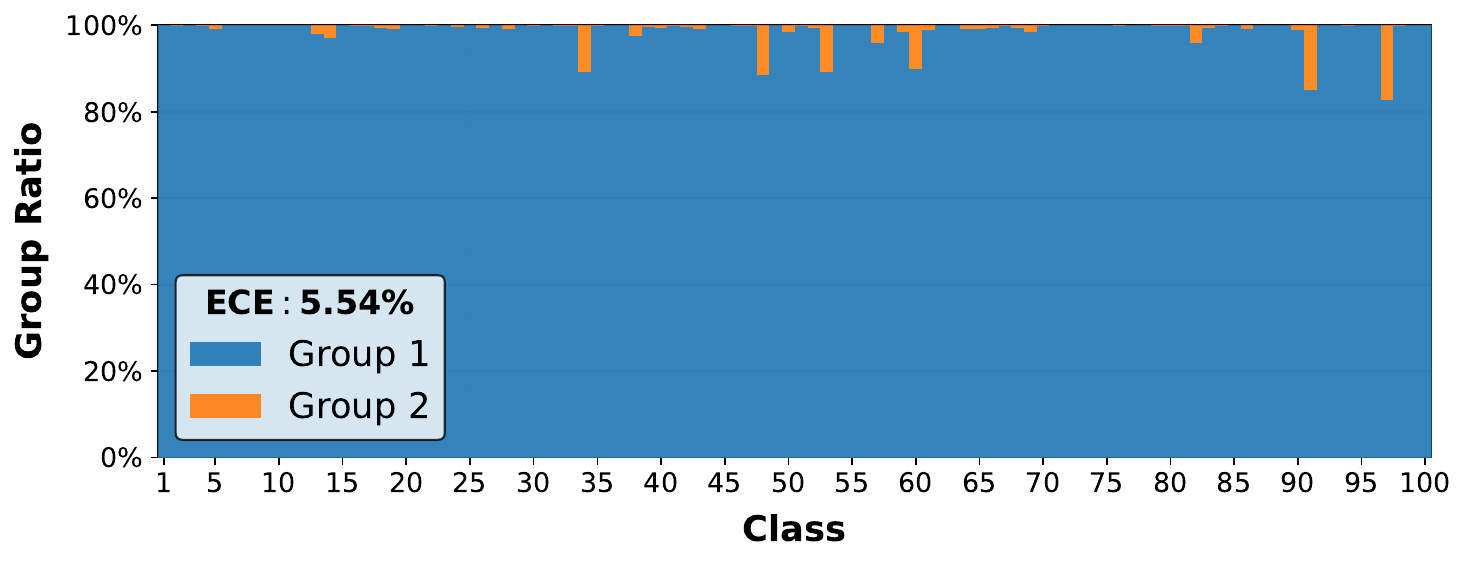}
        \vspace{-4mm}
        \caption{Naive grouping}
        \label{fig:grouping_without_gdca}
    \end{subfigure}
    \hfill
    \begin{subfigure}[H]{0.46\linewidth}
        \centering
        \includegraphics[width=\linewidth]{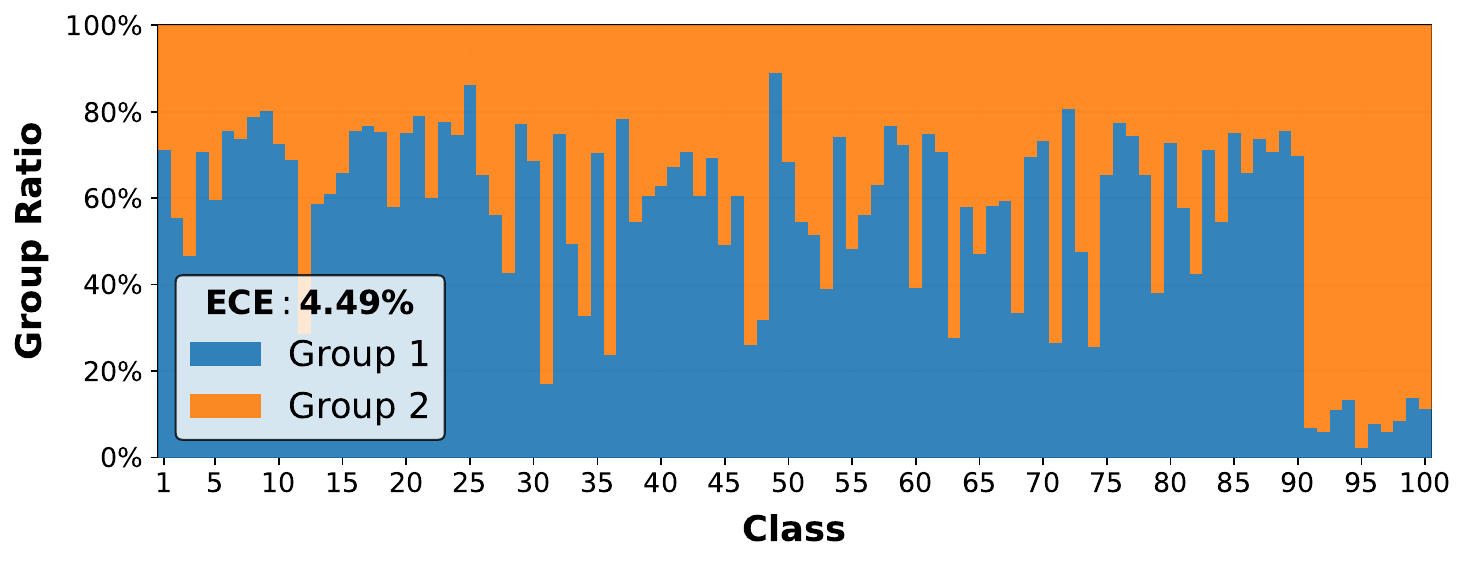}
        \vspace{-4mm}
        \caption{MT-CC grouping}
        \label{fig:MT-CC}
    \end{subfigure}
    \vspace{-2mm}
    \caption{
        Portion of samples from each class allocated to each group. Results are shown after training on the 10th task of CIFAR-100. 
        Compared to 
        (a) naive confidence-based grouping, which yields imbalanced groups,
        (b) MT-CC achieves a more effective group allocation, leading to lower ECE.
    }
    \label{fig:naive_vs_MT-CC}
\end{figure}
\vspace{-1mm}

\textbf{Calibration metrics.}
We evaluate calibration using three standard metrics: Expected Calibration Error (ECE)~\cite{naeini2015obtaining}, Adaptive Expected Calibration Error (AECE)~\cite{mukhoti2020calibrating}, and Static Calibration Error (SCE)~\cite{nixon2020measuringcalibration}. 
ECE measures the gap between predicted confidence and empirical accuracy using fixed confidence bins, while AECE uses adaptive binning based on the confidence distribution of the test data. SCE extends calibration evaluation to the multi-class setting by considering class-wise probabilities.
In our experiments, we use 10 bins for all metrics.

\noindent\textbf{Baselines.} 
We compare MT-CC with representative post-hoc calibration methods.
(i) Vanilla denotes the continual learning model without calibration.
(ii) TS \citep{guo2017calibration} learns a single global temperature.
(iii) ETS \citep{zhang2020mix} extends TS through an ensemble of probability transformations.
(iv) IRM \citep{zhang2020mix} performs non-parametric calibration via multi-class isotonic regression.
(v) PTS \citep{tomani2022parameterized} learns prediction-specific temperatures using a lightweight neural network.
(vi) AdaTS \citep{joy2023sample} estimates instance-wise temperatures using a class-conditional VAE and a lightweight neural network.
(vii) T-CIL \citep{hwang2025t} optimizes TS using adversarially perturbed exemplars for class-incremental learning.
(viii) For stronger baselines, we apply ETS, IRM, PTS, and AdaTS to T-CIL \citep{hwang2025t}, denoted as T-CIL+[method].
To ensure a fair comparison in CIL, we report calibration results under two validation protocols: the raw new-task validation set and the T-CIL-augmented validation set.
Within each protocol, all baselines use the same number of validation samples for temperature tuning. This allows us to compare standard post-hoc methods and CIL-specific methods under consistent validation settings. Importantly, all methods preserve the same accuracy since they operate in a post-hoc manner.

\noindent\textbf{Implementation details.}
For all datasets, the ResNet-based main experiments use a single fully connected layer for the grouping network $g_{\phi}(x)$ with a learning rate of $0.01$. 
% A GDCA regularization weight of $1.0$ is applied across all experiments.
The GDCA regularization weight is set to $1.0$ for ResNet and $0.1$ for ViT experiments.
The initial temperature values are set to $[1.5, 1.5]$, and the validation set is generated using the T-CIL framework and divided into $\mathcal{S}_{\text{grp}}$ and $\mathcal{S}_{\text{cal}}$ with equal sizes. 
The module is optimized using the Adam optimizer \citep{kingma2014adam} with a batch size of $16$ for $50$ epochs. \fix{ The number of groups is set to $M=2$ in our experiments, as the two-group configuration is sufficient to provide reliable calibration performance, as shown in Appendix~\ref{sec:number_of_groups}.} 
The main experiments are conducted based on ER~\citep{chaudhry2019tiny}.
\fix{To further verify the applicability of MT-CC, we additionally compare the performance of different approaches using representative CIL methods.} 
\fix{ViT-based experiments and their implementation details are provided in Appendix~\ref{sec:implementation_details}}.

\begin{table}[t]
\vspace{-2mm}
\centering
\caption{\fix{ImageNet results with varying replay buffer sizes. Since all evaluated methods are post-hoc, they share the same accuracies of the base models: for \textbf{ER}, $32.86$\% (5K), $41.23$\% (10K), $49.52$\% (20K); for \textbf{iCaRL}, $44.08$\% (5K), $49.29$\% (10K), $54.24$\% (20K). Reflecting that SCE values vary with the number of labels, per mille (\textperthousand) is used.}}
\label{tab:imagenet}
\setlength{\tabcolsep}{2pt}
\resizebox{0.875\textwidth}{!}{%
\begin{tabular}{@{}ll|ccc|ccc|ccc@{}}
\toprule
\multicolumn{2}{c|}{\textbf{Replay buffer}} &
\multicolumn{3}{c|}{\textbf{5K}} &
\multicolumn{3}{c|}{\textbf{10K}} &
\multicolumn{3}{c}{\textbf{20K}} \\
\textbf{CL} & \textbf{CA} &
\textbf{ECE($\downarrow$)} &
\textbf{AECE($\downarrow$)} &
\textbf{SCE($\downarrow$)} &
\textbf{ECE($\downarrow$)} &
\textbf{AECE($\downarrow$)} &
\textbf{SCE($\downarrow$)} &
\textbf{ECE($\downarrow$)} &
\textbf{AECE($\downarrow$)} &
\textbf{SCE($\downarrow$)} \\
\midrule
\midrule

\textbf{ER}
& Vanilla
& \meanstdsmall{29.95}{0.30}
& \meanstdsmall{29.95}{0.30}
& \meanstdsmall{2.18}{0.02}
& \meanstdsmall{22.46}{0.19}
& \meanstdsmall{22.46}{0.19}
& \meanstdsmall{1.80}{0.02}
& \meanstdsmall{16.39}{0.23}
& \meanstdsmall{16.38}{0.23}
& \meanstdsmall{1.45}{0.02} \\

& TS \cite{guo2017calibration}
& \meanstdsmall{29.66}{0.33}
& \meanstdsmall{29.63}{0.33}
& \meanstdsmall{2.16}{0.02}
& \meanstdsmall{22.67}{0.17}
& \meanstdsmall{22.65}{0.18}
& \meanstdsmall{1.79}{0.02}
& \meanstdsmall{16.83}{0.18}
& \meanstdsmall{16.80}{0.20}
& \meanstdsmall{1.44}{0.01} \\

& ETS \cite{zhang2020mix}
& \meanstdsmall{30.68}{0.25}
& \meanstdsmall{30.65}{0.26}
& \meanstdsmall{1.09}{0.01}
& \meanstdsmall{23.53}{0.16}
& \meanstdsmall{23.53}{0.16}
& \meanstdsmall{1.09}{0.01}
& \meanstdsmall{17.62}{0.22}
& \meanstdsmall{17.58}{0.23}
& \meanstdsmall{0.91}{0.00} \\

& PTS \cite{tomani2022parameterized}
& \meanstdsmall{30.32}{0.44}
& \meanstdsmall{30.32}{0.44}
& \meanstdsmall{1.09}{0.01}
& \meanstdsmall{22.79}{0.36}
& \meanstdsmall{22.80}{0.36}
& \meanstdsmall{1.89}{0.01}
& \meanstdsmall{16.52}{0.27}
& \meanstdsmall{16.51}{0.28}
& \meanstdsmall{0.91}{0.01} \\

& T-CIL \cite{hwang2025t}
& \meanstdsmall{13.45}{0.15}
& \meanstdsmall{13.43}{0.13}
& \meanstdsmall{1.79}{0.01}
& \meanstdsmall{7.74}{0.23}
& \meanstdsmall{7.72}{0.24}
& \meanstdsmall{1.49}{0.01}
& \meanstdsmall{10.57}{0.22}
& \meanstdsmall{10.52}{0.24}
& \meanstdsmall{1.27}{0.00} \\

\rowcolor{blue!15}
& \textbf{MT-CC}
& \boldmath\meanstdsmall{12.59}{0.56}
& \boldmath\meanstdsmall{12.62}{0.56}
& \boldmath\meanstdsmall{1.05}{0.01}
& \boldmath\meanstdsmall{7.25}{0.56}
& \boldmath\meanstdsmall{7.26}{0.56}
& \boldmath\meanstdsmall{0.91}{0.01}
& \boldmath\meanstdsmall{8.44}{0.69}
& \boldmath\meanstdsmall{8.44}{0.69}
& \boldmath\meanstdsmall{0.82}{0.02} \\

\midrule

\textbf{iCaRL}
& Vanilla
& \meanstdsmall{22.85}{0.18}
& \meanstdsmall{22.85}{0.19}
& \meanstdsmall{1.93}{0.02}
& \meanstdsmall{17.91}{0.23}
& \meanstdsmall{17.90}{0.23}
& \meanstdsmall{1.68}{0.02}
& \meanstdsmall{12.94}{0.83}
& \meanstdsmall{12.93}{0.82}
& \meanstdsmall{1.44}{0.05} \\

& TS \cite{guo2017calibration}
& \meanstdsmall{22.24}{0.31}
& \meanstdsmall{22.22}{0.29}
& \meanstdsmall{1.93}{0.02}
& \meanstdsmall{18.08}{0.38}
& \meanstdsmall{18.07}{0.37}
& \meanstdsmall{1.68}{0.02}
& \meanstdsmall{16.18}{2.88}
& \meanstdsmall{16.16}{2.88}
& \meanstdsmall{1.47}{0.10} \\

& ETS \cite{zhang2020mix}
& \meanstdsmall{22.79}{0.12}
& \meanstdsmall{22.78}{0.12}
& \meanstdsmall{1.16}{0.01}
& \meanstdsmall{18.67}{0.43}
& \meanstdsmall{18.64}{0.44}
& \meanstdsmall{1.03}{0.01}
& \meanstdsmall{15.10}{0.11}
& \meanstdsmall{15.08}{0.11}
& \meanstdsmall{0.90}{0.01} \\

& PTS \cite{tomani2022parameterized}
& \meanstdsmall{21.82}{0.05}
& \meanstdsmall{21.83}{0.04}
& \boldmath\meanstdsmall{0.96}{0.00}
& \meanstdsmall{17.63}{0.16}
& \meanstdsmall{17.63}{0.17}
& \boldmath\meanstdsmall{0.83}{0.01}
& \meanstdsmall{14.07}{0.09}
& \meanstdsmall{14.09}{0.10}
& \meanstdsmall{0.99}{0.01} \\

& T-CIL \cite{hwang2025t}
& \meanstdsmall{5.99}{0.30}
& \meanstdsmall{5.98}{0.31}
& \meanstdsmall{1.80}{0.02}
& \meanstdsmall{6.84}{0.01}
& \meanstdsmall{6.80}{0.04}
& \meanstdsmall{1.57}{0.01}
& \meanstdsmall{9.69}{1.48}
& \meanstdsmall{9.64}{1.47}
& \meanstdsmall{1.36}{0.03} \\

\rowcolor{blue!15}
& \textbf{MT-CC}
& \boldmath\meanstdsmall{4.87}{0.26}
& \boldmath\meanstdsmall{4.88}{0.25}
& \meanstdsmall{1.05}{0.01}
& \boldmath\meanstdsmall{5.16}{0.14}
& \boldmath\meanstdsmall{5.12}{0.12}
& \meanstdsmall{0.95}{0.01}
& \boldmath\meanstdsmall{6.88}{0.79}
& \boldmath\meanstdsmall{6.80}{0.79}
& \boldmath\meanstdsmall{0.85}{0.01} \\

\bottomrule
\end{tabular}%
}
\end{table}

\begin{table}[t]
\vspace{-2mm}
\centering
\setlength{\tabcolsep}{2pt}
\caption{
ViT-based CIL post-hoc calibration results over five seeds. Base accuracy is preserved at 90.99\% on CIFAR-100 and 79.23\% on ImageNet-R. Metrics are in $10^{-2}$ scale.
}
\label{tab:vit_cil_results}
\begin{adjustbox}{max width=0.575\columnwidth}
\begin{tabular}{@{}l|ccc|ccc@{}}
\toprule
\multirow{2}{*}{\textbf{Method}}
& \multicolumn{3}{c|}{\textbf{CIFAR-100}}
& \multicolumn{3}{c}{\textbf{ImageNet-R}} \\
\cmidrule(lr){2-4} \cmidrule(lr){5-7}
& \textbf{ECE} & \textbf{AECE} & \textbf{SCE}
& \textbf{ECE} & \textbf{AECE} & \textbf{SCE} \\
\midrule
Vanilla
& \meanstdsmall{6.23}{0.19}
& \meanstdsmall{0.41}{0.02}
& \meanstdsmall{0.34}{0.02}
& \meanstdsmall{10.68}{0.31}
& \meanstdsmall{0.38}{0.01}
& \meanstdsmall{0.37}{0.03} \\

TS \citep{guo2017calibration}
& \meanstdsmall{6.44}{0.20}
& \meanstdsmall{0.30}{0.02}
& \meanstdsmall{0.34}{0.03}
& \meanstdsmall{11.80}{0.29}
& \meanstdsmall{0.32}{0.01}
& \meanstdsmall{0.38}{0.03} \\

ETS \citep{zhang2020mix}
& \meanstdsmall{9.82}{0.37}
& \meanstdsmall{0.29}{0.02}
& \meanstdsmall{0.34}{0.03}
& \meanstdsmall{17.30}{0.93}
& \meanstdsmall{0.29}{0.02}
& \meanstdsmall{0.39}{0.03}\\

IRM \citep{zhang2020mix}
& \meanstdsmall{8.39}{0.34}
& \meanstdsmall{0.31}{0.01}
& \meanstdsmall{0.36}{0.03}
& \meanstdsmall{15.51}{0.99}
& \meanstdsmall{0.28}{0.02}
& \meanstdsmall{0.41}{0.03} \\

PTS \citep{tomani2022parameterized}
& \meanstdsmall{9.66}{0.44}
& \meanstdsmall{0.29}{0.02}
& \meanstdsmall{0.34}{0.03}
& \meanstdsmall{18.43}{1.03}
& \meanstdsmall{0.27}{0.02}
& \meanstdsmall{0.38}{0.03} \\

AdaTS \citep{joy2023sample} 
& \meanstdsmall{6.00}{0.68}
& \meanstdsmall{0.28}{0.02}
& \meanstdsmall{0.34}{0.02}
& \meanstdsmall{8.93}{0.66}
& \meanstdsmall{0.28}{0.01}
& \meanstdsmall{0.37}{0.02} \\

\rowcolor{blue!15} \textbf{MT-CC w/o aug}
& \boldmath\meanstdsmall{5.10}{0.49}
& \boldmath\meanstdsmall{0.27}{0.01}
& \boldmath\meanstdsmall{0.33}{0.02}
& \boldmath\meanstdsmall{7.99}{0.63}
& \boldmath\meanstdsmall{0.28}{0.02}
& \boldmath\meanstdsmall{0.36}{0.03} \\

\midrule
T-CIL \citep{hwang2025t}
& \meanstdsmall{5.23}{0.54}
& \meanstdsmall{0.30}{0.02}
& \meanstdsmall{0.34}{0.03}
& \meanstdsmall{8.04}{0.71}
& \meanstdsmall{0.29}{0.02}
& \meanstdsmall{0.36}{0.03} \\

T-CIL + ETS 
& \meanstdsmall{5.49}{0.53}
& \meanstdsmall{0.29}{0.01}
& \meanstdsmall{0.35}{0.03}
& \meanstdsmall{8.13}{0.72}
& \meanstdsmall{0.28}{0.01}
& \meanstdsmall{0.37}{0.02} \\

T-CIL + IRM 
& \meanstdsmall{5.51}{0.57}
& \meanstdsmall{0.29}{0.02}
& \meanstdsmall{0.34}{0.03}
& \meanstdsmall{7.95}{0.64}
& \meanstdsmall{0.28}{0.01}
& \meanstdsmall{0.37}{0.03} \\

T-CIL + PTS
& \meanstdsmall{6.03}{0.66}
& \meanstdsmall{0.29}{0.02}
& \meanstdsmall{0.34}{0.03}
& \meanstdsmall{10.00}{0.84}
& \meanstdsmall{0.29}{0.01}
& \meanstdsmall{0.38}{0.02} \\

T-CIL + AdaTS 
& \meanstdsmall{6.04}{0.61}
& \meanstdsmall{0.27}{0.02}
& \meanstdsmall{0.33}{0.02}
& \meanstdsmall{9.03}{0.85}
& \meanstdsmall{0.28}{0.01}
& \meanstdsmall{0.37}{0.02} \\

\rowcolor{blue!15}
\textbf{MT-CC}
& \boldmath\meanstdsmall{4.57}{0.28}
& \boldmath\meanstdsmall{0.26}{0.02}
& \boldmath\meanstdsmall{0.33}{0.03}
& \boldmath\meanstdsmall{7.55}{0.29}
& \boldmath\meanstdsmall{0.26}{0.01}
& \boldmath\meanstdsmall{0.30}{0.03} \\
\bottomrule
\end{tabular}
\end{adjustbox}
\vspace{-3mm}
\end{table}

\subsection{Main Experimental Results}
\label{sec:main_results}
\vspace{-1mm}

\textbf{Comparison with baselines.}
Table~\ref{tab:main_results} compares MT-CC with representative accuracy-preserving post-hoc calibration methods.
MT-CC achieves the lowest ECE and AECE across CIFAR-10, CIFAR-100, and Tiny-ImageNet, while remaining competitive in SCE.
We further report \emph{MT-CC w/o aug}, which denotes MT-CC trained on the raw new-task validation set without T-CIL's augmented validation samples. It remains competitive among raw-validation baselines, suggesting that the multi-group temperature design drives the calibration gains.

\noindent\textbf{Analysis of group partitioning behavior.} Figure~\ref{fig:naive_vs_MT-CC} compares the group allocation results of naive grouping and MT-CC. Naive grouping, which only takes confidence values as input, leads to a highly imbalanced class-group distribution in the final task.  
MT-CC maintains a balanced distribution across groups and effectively distinguishes samples with different calibration behaviors. The grouping network in MT-CC leads to a higher temperature in Group~2, which contains not only the majority of new-task samples (classes 91 -- 100) but also many of the old-task samples (classes 1 -- 90). 
As a result, this adaptive grouping and temperature assignment lead to improved calibration performance.

\begin{table}[t]
\centering
\caption{Ablation on grouping and input signals of MT-CC.}
\label{tab:ablation_grouping_inputs}
\begin{adjustbox}{max width=0.95\textwidth}
\begin{tabular}{l|ccc|cc|cc|cc}
\toprule
& \multicolumn{3}{c|}{\textbf{Ablated Components}} 
& \multicolumn{2}{c|}{\textbf{CIFAR-10}} 
& \multicolumn{2}{c|}{\textbf{CIFAR-100}} 
& \multicolumn{2}{c}{\textbf{Tiny-ImageNet}} \\
\cmidrule(lr){2-4} \cmidrule(lr){5-10}
\textbf{Exp} 
& \textbf{Logit Margin} & \textbf{Old/New Confidence} & \textbf{GDCA}
& \textbf{ECE ($\downarrow$)} & \textbf{AECE ($\downarrow$)} 
& \textbf{ECE ($\downarrow$)} & \textbf{AECE ($\downarrow$)}
& \textbf{ECE ($\downarrow$)} & \textbf{AECE ($\downarrow$)} \\
\midrule
$\mathcal{A}$ (All)      & $\checkmark$ & $\checkmark$ & $\checkmark$ & \meanstdsmall{10.18}{1.48} & \meanstdsmall{10.16}{1.54} & \boldmath\meanstdsmall{4.49}{0.53} & \boldmath\meanstdsmall{4.44}{0.49} & \boldmath\meanstdsmall{15.82}{0.29} & \boldmath\meanstdsmall{15.80}{0.32} \\
$\mathcal{B}$   & $\times$     & $\checkmark$ & $\checkmark$ & \boldmath\meanstdsmall{10.12}{1.52}  & \boldmath\meanstdsmall{10.12}{1.53} & \meanstdsmall{4.95}{0.37} & \meanstdsmall{4.93}{0.34} &\meanstdsmall{15.90}{0.36}  & \meanstdsmall{15.90}{0.41} \\
$\mathcal{C}$   & $\checkmark$ & $\checkmark$ & $\times$ & \meanstdsmall{11.81}{1.41} & \meanstdsmall{11.81}{1.40} & \meanstdsmall{4.78}{0.42} & \meanstdsmall{4.79}{0.40} & \meanstdsmall{15.93}{0.37} & \meanstdsmall{15.94}{0.42} \\
$\mathcal{D}$   & $\checkmark$ & $\times$ & $\checkmark$     & \meanstdsmall{10.91}{1.26} & \meanstdsmall{10.91}{1.26} & \meanstdsmall{4.63}{0.30} & \meanstdsmall{4.65}{0.26} & \meanstdsmall{19.48}{0.40} & \meanstdsmall{19.47}{0.44} \\
$\mathcal{E}$   & $\times$ & $\times$     & $\times$     & \meanstdsmall{12.45}{1.23} & \meanstdsmall{12.42}{1.20} & \meanstdsmall{5.54}{0.42} &  \meanstdsmall{5.54}{0.41}& \meanstdsmall{19.69}{0.83} & \meanstdsmall{19.68}{0.86} \\
\bottomrule
\end{tabular}
\end{adjustbox}
\vspace{-3mm}
\end{table}

\vspace{-1mm}
\subsection{Further Studies}
\vspace{-1mm}
\noindent\textbf{Scalability and backbone generality (ImageNet and ViT).}
We further evaluate MT-CC on ImageNet and ViT-based CIL settings. 
Table~\ref{tab:imagenet} shows that MT-CC remains effective on ImageNet with 1{,}000 classes across different memory budgets and replay-based CIL methods.
Table~\ref{tab:vit_cil_results} further demonstrates that MT-CC generalizes beyond CNN backbones, achieving strong calibration performance with ViT on CIFAR-100 and ImageNet-R while preserving the same base accuracy.
Overall, these results indicate that MT-CC is robust to larger output spaces and different backbone architectures.

\begin{figure}[t]
\centering
\begin{minipage}{0.5\textwidth}
    \centering
    \captionsetup{type=figure}
    \includegraphics[width=\linewidth]{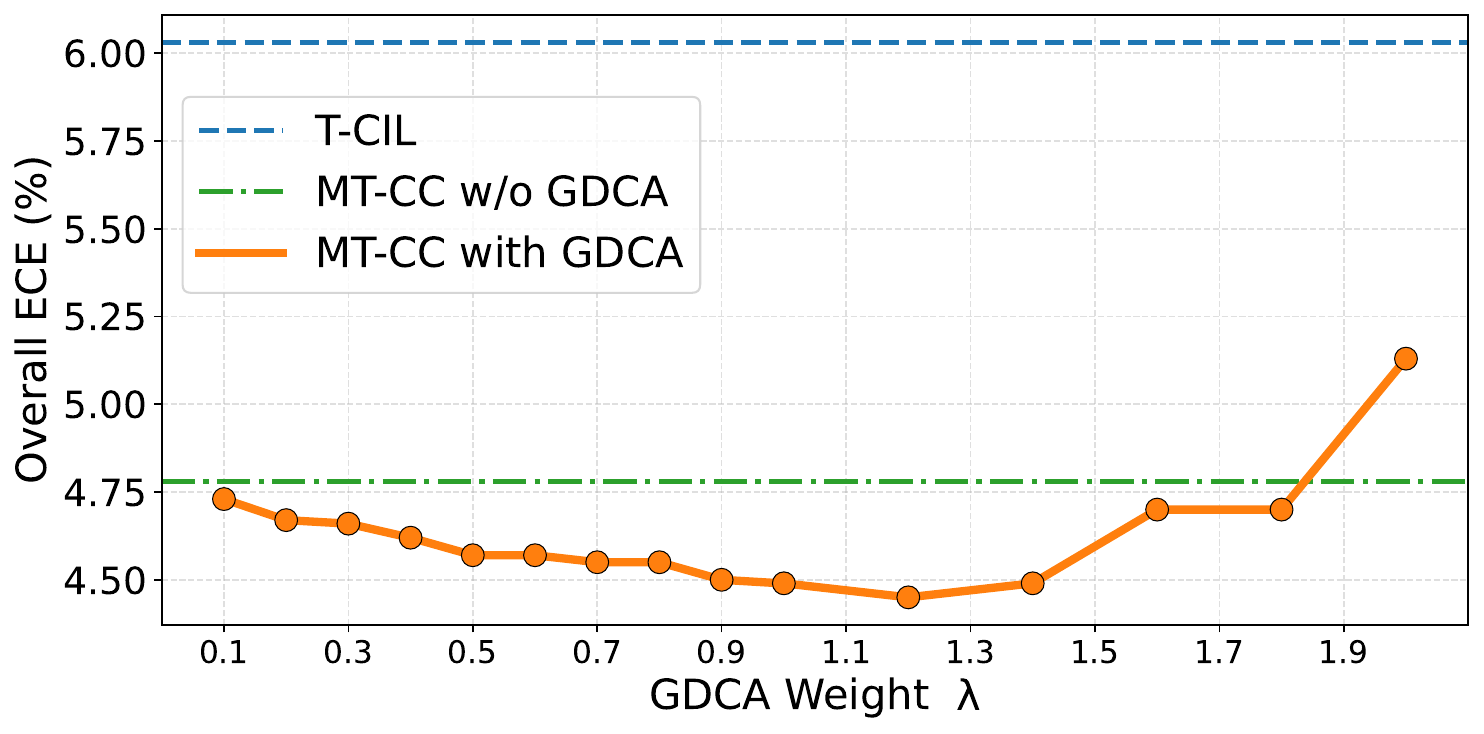}
    \caption{Effect of GDCA weight $\lambda$ on overall calibration. Across a broad range of $\lambda$ values, both MT-CC variants (with and without GDCA) consistently outperform the baseline.}
    \label{fig:gdca_weight_vs_ece}
\end{minipage}
\hspace{0.05\textwidth}
\begin{minipage}{0.43\textwidth}
    \footnotesize
    \centering
    \captionsetup{type=table, font=footnotesize}
    \captionof{table}{Versatility with other CIL methods.}
    \label{tab:cil_compat}
    \vspace{-2mm}
    \resizebox{\linewidth}{!}{
    \begin{tabular}{lccc}
    \toprule
    \textbf{Method} & \textbf{ECE} & \textbf{AECE} & \textbf{SCE} \\
    \midrule
    EEIL + TS       & \meanstdsmall{14.29}{0.63} & \meanstdsmall{14.29}{0.64} & \meanstdsmall{1.05}{0.02} \\
    EEIL + T-CIL    & \meanstdsmall{4.28}{0.35}  & \meanstdsmall{4.34}{0.28}  & \meanstdsmall{1.01}{0.03} \\
    \rowcolor{blue!15} EEIL + MT-CC    & \boldmath\meanstdsmall{3.67}{0.20}  & \boldmath\meanstdsmall{3.71}{0.18}  & \boldmath\meanstdsmall{0.58}{0.01} \\
    \midrule
    BiC  + TS       & \meanstdsmall{34.19}{1.54} & \meanstdsmall{34.13}{1.55} & \meanstdsmall{1.55}{0.05} \\
    BiC  + T-CIL    & \meanstdsmall{27.52}{0.85} & \meanstdsmall{27.44}{0.83} & \meanstdsmall{1.41}{0.04} \\
    \rowcolor{blue!15} BiC  + MT-CC    & \boldmath\meanstdsmall{26.73}{0.99} & \boldmath\meanstdsmall{26.65}{0.97} & \boldmath\meanstdsmall{0.84}{0.03} \\
    \midrule
    iCaRL + TS      & \meanstdsmall{17.89}{0.41} & \meanstdsmall{17.83}{0.42} & \meanstdsmall{1.24}{0.03} \\
    iCaRL + T-CIL   & \meanstdsmall{8.64}{1.12}  & \meanstdsmall{8.62}{1.14}  & \meanstdsmall{1.13}{0.04} \\
    \rowcolor{blue!15} iCaRL + MT-CC   & \boldmath\meanstdsmall{6.00}{1.02}  & \boldmath\meanstdsmall{6.10}{0.99}  & \boldmath\meanstdsmall{0.68}{0.02} \\
    \midrule
    IL2M + TS       & \meanstdsmall{10.15}{0.43} & \meanstdsmall{10.10}{0.41} & \meanstdsmall{0.88}{0.02} \\
    IL2M + T-CIL    & \meanstdsmall{8.54}{1.37}  & \meanstdsmall{8.62}{1.33}  & \meanstdsmall{0.90}{0.02} \\
    \rowcolor{blue!15} IL2M + MT-CC    & \boldmath\meanstdsmall{7.72}{0.89}  & \boldmath\meanstdsmall{7.75}{0.86}  & \boldmath\meanstdsmall{0.47}{0.02} \\
    \midrule
    DER + TS        & \meanstdsmall{17.68}{0.18} & \meanstdsmall{17.63}{0.20} & \meanstdsmall{0.90}{0.01} \\
    DER + T-CIL     & \meanstdsmall{6.18}{0.31}  & \meanstdsmall{6.12}{0.30}  & \meanstdsmall{0.79}{0.02} \\
    \rowcolor{blue!15} DER + MT-CC     & \boldmath\meanstdsmall{4.10}{0.39} & \boldmath\meanstdsmall{4.06}{0.41} & \boldmath\meanstdsmall{0.69}{0.01} \\
    \bottomrule
    \end{tabular}}
\end{minipage}
\end{figure}

\begin{table}[!t]
\vspace{-4mm}
\centering
\caption{Comparison of task-wise grouping, per-task TS, and the proposed MT-CC.}
\label{tab:ece_cifar100}
\begin{adjustbox}{max width=\textwidth}
\begin{tabular}{lccccccccccc}
\toprule
\textbf{Method} & \textbf{T1} & \textbf{T2} & \textbf{T3} & \textbf{T4} & \textbf{T5} & \textbf{T6} & \textbf{T7} & \textbf{T8} & \textbf{T9} & \textbf{T10} & \textbf{Overall} \\
\midrule
GT Old/New TS
& \boldmath\meanstdsmall{2.18}{0.63}
& \meanstdsmall{11.50}{1.77}
& \meanstdsmall{5.43}{2.05}
& \meanstdsmall{4.81}{1.11}
& \meanstdsmall{3.47}{1.28}
& \meanstdsmall{3.76}{0.46}
& \meanstdsmall{4.35}{1.97}
& \meanstdsmall{6.00}{2.34}
& \meanstdsmall{7.88}{2.22}
& \meanstdsmall{10.37}{2.02}
& \meanstdsmall{5.97}{0.32} \\

GT Per-task TS
& \boldmath\meanstdsmall{2.18}{0.63}
& \meanstdsmall{10.79}{1.60}
& \meanstdsmall{5.42}{2.03}
& \meanstdsmall{4.80}{1.11}
& \meanstdsmall{3.47}{1.27}
& \meanstdsmall{3.77}{0.46}
& \boldmath\meanstdsmall{4.34}{1.96}
& \meanstdsmall{6.00}{2.33}
& \meanstdsmall{7.88}{2.22}
& \meanstdsmall{10.42}{1.93}
& \meanstdsmall{5.91}{0.31} \\

\rowcolor{blue!15} MT-CC 
& \boldmath\meanstdsmall{2.18}{0.63}
& \boldmath\meanstdsmall{6.99}{2.03}
& \boldmath\meanstdsmall{3.22}{0.94}
& \boldmath\meanstdsmall{2.32}{0.50}
& \boldmath\meanstdsmall{3.16}{0.65}
& \boldmath\meanstdsmall{2.49}{0.36}
& \meanstdsmall{4.75}{2.39}
& \boldmath\meanstdsmall{4.72}{3.27}
& \boldmath\meanstdsmall{7.11}{2.75}
& \boldmath\meanstdsmall{7.97}{2.13}
& \boldmath\meanstdsmall{4.49}{0.53} \\
\bottomrule
\end{tabular}
\end{adjustbox}
\vspace{-3mm}
\end{table}

\noindent\textbf{Ablation on grouping and input signals.}
Table~\ref{tab:ablation_grouping_inputs} analyzes the contribution of MT-CC components.
Removing either the logit margin or old/new confidence feature generally increases ECE and AECE, indicating that both contribute useful calibration signals.
Eliminating GDCA further degrades performance, confirming its role in improving groupwise calibration.

\noindent\textbf{Effect of GDCA regularization.} Figure~\ref{fig:gdca_weight_vs_ece} illustrates the effect of the GDCA weight, 
%$\lambda$ in Eq.~\ref{eq:gdca_total}, 
evaluated on the CIFAR-100 dataset. GDCA improves ECE across a wide range of $\lambda$ values, confirming its effectiveness. Notably, both MT-CC variants, with and without GDCA, consistently outperform the T-CIL baseline, further demonstrating the robustness of our method. 
%As shown in Figure~\ref{fig:gdca_weight_vs_ece}, increasing the GDCA weight initially improves calibration, reducing overall ECE until it saturates around weight = 1.2.  
%This shows that mild regularization promotes groupwise consistency without over-smoothing temperature parameters.  

\noindent\textbf{Compatibility with other CIL methods.} 
To further evaluate the compatibility of MT-CC, we conduct experiments by integrating it with several milestone replay-based CIL methods, including EEIL~\citep{castro2018end}, iCaRL~\citep{rebuffi2017icarl}, BiC~\citep{wu2019large}, IL2M~\citep{Belouadah_2019_ICCV}, and DER~\citep{yan2021dynamically}.
CIFAR-100 is adopted for experiments. As shown in Table~\ref{tab:cil_compat}, MT-CC consistently achieves the best calibration performance across all frameworks, demonstrating its plug-and-play applicability to diverse CIL methods.

\noindent\textbf{Comparison with ground-truth grouping baselines.} 
Table~\ref{tab:ece_cifar100} compares MT-CC with two oracle ground-truth grouping baselines on CIFAR-100.
The first baseline, \emph{GT Old/New TS}, splits samples into two groups, old-task classes $\mathcal{Y}_{1:b-1}$ and the new-task classes $\mathcal{Y}_{b}$, and applies temperature scaling to each group.
The second baseline, \emph{GT Per-task TS}, uses a more fine-grained oracle partition by assigning samples to their corresponding task groups and learning a separate temperature for each task.
Since each per-task group contains fewer calibration samples, we use a smaller learning rate for GT Per-task TS to ensure stable and fair optimization.
Despite this oracle advantage, MT-CC achieves lower ECE than both grouping baselines in most tasks.
These results show that MT-CC learns a more effective calibration boundary than rule-based task grouping.

\section{Conclusion}
In this work, we proposed MT-CC, a multi-group temperature scaling strategy for handling asymmetric calibration behavior in CIL. MT-CC incorporates a strategically designed grouping network that leverages old/new-task confidence and logit margin as inputs. The GDCA loss further stabilizes this network and promotes intra-group consistency without sacrificing accuracy. Extensive experiments demonstrate that MT-CC achieves state-of-the-art calibration performance in CIL, providing a promising direction for building reliable models in continual learning settings. 

\textbf{Limitations and future work.} A deeper theoretical understanding of MT-CC remains to be developed. Extending MT-CC to different continual learning domains (e.g., large language models) is also an important direction for future work. 

% \subsection*{AI Use Statement}

% Generative AI tools were used to assist with literature exploration,
% research brainstorming, feedback on experimental design, code development
% and debugging, interpretation and presentation of experimental results,
% and manuscript writing and editing. All AI-assisted suggestions, code,
% analyses, and text were independently reviewed and verified by the authors.
% The authors take full responsibility for the scientific claims, experimental
% results, and final content of this work.

% \subsection*{Ethics statement}

% This work uses publicly available benchmark datasets and does not involve human subjects or personally identifiable information. We do not identify any specific ethical concerns beyond those generally associated with the datasets and continual learning models used in our experiments.

% \subsection*{Reproducibility statement}

% We provide the methodological and experimental details required to reproduce our results throughout the main paper and appendix. The proposed MT-CC framework and its optimization are described in the main text, while implementation details, hyperparameters, additional ablations, and experimental settings are provided in the appendix. We additionally provide the source code as supplementary material to facilitate reproducibility. All reported results are averaged over multiple random seeds.

\subsubsection*{Acknowledgments}
This work was supported in part by the National Research Foundation of Korea (NRF) grant funded by the Korea government (MSIT) (RS-2026-25470983), the Institute of Information \& Communications Technology Planning \& Evaluation (IITP) grant funded by the Korea government (MSIT) (RS-2024-00457882, AI Research Hub Project, IITP-2026-RS-2026-25544188, Leading Generative AI Human Resources Development), and the Technology Development Program of MSS (RS-2025-25464351). Dong-Jun Han is the corresponding author.

%This work was supported in part by the National Research Foundation of Korea (NRF) grant funded by the Korea government (MSIT) (RS-2026-25470983), the Institute of Information \& Communications Technology Planning \& Evaluation (IITP) grant funded by the Korea government (MSIT) (RS-2024-00457882, AI Research Hub Project), and the Technology Development Program of MSS (RS-2025-25464351). Dong-Jun Han is the corresponding author.

\bibliographystyle{plainnat}
\bibliography{references}

\clearpage
% Start a separate TOC for the appendix.
\startcontents[appendix]
\appendix
\begin{center}
{\Large\bfseries Appendix}\\[2mm]
\end{center}

% Print only appendix sections/subsections.
{\hypersetup{linkcolor=black}
\printcontents[appendix]{}{1}{
    \setcounter{tocdepth}{2}
}}

\vspace{3mm}
\clearpage

\begin{table}[t]
\centering
\caption{Hyperparameters for Group Calibration Experiments}
\label{tab:hyperparameters}

\begin{adjustbox}{max width=\textwidth}
\begin{tabular}{lcccccc}
\toprule
& \multicolumn{4}{c}{\textbf{ResNet-based}} 
& \multicolumn{2}{c}{\textbf{ViT-based}} \\
\cmidrule(lr){2-5} \cmidrule(lr){6-7}
\textbf{Hyperparameter} 
& \textbf{CIFAR-10} 
& \textbf{CIFAR-100} 
& \textbf{Tiny-ImageNet} 
& \textbf{ImageNet} 
& \textbf{CIFAR-100} 
& \textbf{ImageNet-R} \\
\midrule
\multicolumn{7}{c}{\textit{Dataset Configuration}} \\
\midrule
Dataset & cifar10 & cifar100 & tiny-imagenet & imagenet & cifar100 & ImageNet-R \\
Number of Classes & 10 & 100 & 200 & 1000 & 100 & 200 \\
New Classes per Task & 2 & 10 & 20 & 100 & 10 & 20 \\
Total Tasks & 5 & 10 & 10 & 10 & 10 & 10 \\
Batch Size & 128 & 128 & 128 & 256 & 128 & 128 \\
Validation Size  & 500 & 500 & 100 & 5000 & 500 & 500 \\
Replay Buffer Size & 0.2K & 2K & 2K & 5K/10K/20K & 2K & 2K \\
\midrule
\multicolumn{7}{c}{\textit{Model Architecture}} \\
\midrule
Model & ResNet-32 & ResNet-32 & ResNet-18 & ResNet-18 & ViT-B/16 & ViT-B/16 \\
Image Size & 32 & 32 & 64 & 224 & 224 & 224 \\
\midrule
\multicolumn{7}{c}{\textit{Temperature Scaling}} \\
\midrule
Calibration Epochs & 100 & 100 & 100 & 100 & 50 & 50 \\
Calibration Batch Size & 128 & 128 & 128 & 128 & 128 & 128 \\
Temp Learning Rate & 0.1 & 0.1 & 0.1 & 0.1 & 0.03 & 0.03 \\
\midrule
\multicolumn{7}{c}{\textit{MT-CC}} \\
\midrule
Number of Groups & 2 & 2 & 2 & 2 & 2 & 2 \\
Group Data Ratio & 0.5 & 0.5 & 0.5 & 0.5 & 0.5 & 0.5 \\
\midrule
\multicolumn{7}{c}{\textit{Grouping Network}} \\
\midrule
Optimizer & Adam & Adam & Adam & Adam & Adam & Adam \\
Learning Rate & 0.01 & 0.01 & 0.01 & 0.01 & 0.001 & 0.001 \\
Epochs & 50 & 50 & 50 & 50 & 50 & 50 \\
% Grouping Network Layers & 1  & 1  & 1 & 1 & 1 & 1 \\
\midrule
\multicolumn{7}{c}{\textit{Temperature Initialization}} \\
\midrule
Init Values & [1.5, 1.5] & [1.5, 1.5] & [1.5, 1.5] & [1.5, 1.5] & [1.0, 1.0] & [1.0, 1.0] \\
\midrule
\multicolumn{7}{c}{\textit{Groupwise DCA}} \\
\midrule
GDCA Weight & 1.0 & 1.0 & 1.0 & 1.0 & 0.1 & 0.1 \\
\bottomrule
\end{tabular}
\end{adjustbox}

\end{table}

\section{Training Hyperparameters and Implementation Details}
\label{sec:implementation_details}
\vspace{-3mm}
\paragraph{Implementation details.} All calibration hyperparameters used for MT-CC and the corresponding baselines are summarized in Table~\ref{tab:hyperparameters}. Calibration is performed post-hoc and independently of backbone training. For the underlying CIL methods, we follow their original training protocols unless otherwise specified. For ER, CIFAR-10 and CIFAR-100 are trained for 200 epochs with an initial learning rate of 0.1 and step decay at epochs 100 and 150, using momentum 0.9 and an $\ell_2$ weight decay of $2\times10^{-4}$. Tiny-ImageNet follows the same optimizer and learning-rate schedule with a batch size of 256. Following the original DER protocol, DER additionally uses a 50-epoch decoupled stage with an initial learning rate of 0.05 and decay at epochs 15 and 30, together with a distillation temperature of 5.0. Its pretraining stage is run for 200 epochs with an initial learning rate of 0.1 and decay at epochs 60, 120, and 160. For the remaining CIL baselines, we follow the hyperparameters reported in their original publications.
Before being provided to the grouping network, the logit margin is min-max normalized to ensure a comparable input scale with the other calibration features.
To ensure positive temperature values and stable optimization, each group-specific temperature is parameterized in log-space. For each group $m\in\{1,\ldots,M\}$, we introduce an unconstrained parameter $\tau_m\in\mathbb{R}$ and define
$
T_m=\exp(\tau_m).
$
Accordingly,
$
\frac{\partial \mathcal{L}}{\partial \tau_m}
=
\frac{\partial \mathcal{L}}{\partial T_m}T_m.
$
At the first task, $\tau_m$ is initialized from the predefined initial temperature, while at subsequent tasks it is initialized from the optimized temperature obtained at the previous task.

\vspace{-3mm}
\paragraph{Grouping network input dimensionality.}
The grouping network takes the full softmax probability vector $\mathbf{p}$ as part of its input. To keep its input dimensionality fixed throughout continual learning, we represent $\mathbf{p}$ in the full class space and zero-pad the entries corresponding to unseen classes. Therefore, although the number of encountered classes increases over tasks, the dimensionality of the grouping-network input remains unchanged. Since the grouping network consists of only a lightweight fully connected layer, this fixed-dimensional representation introduces negligible additional overhead.

\subsection{Representative Replay-Based CIL Methods}

We provide brief descriptions of the replay-based CIL methods evaluated in our experiments.

\textbf{EEIL} performs end-to-end incremental learning by jointly optimizing the classification loss and a knowledge distillation objective over replayed exemplars, mitigating feature drift across tasks.

\textbf{iCaRL} combines exemplar replay with knowledge distillation and a nearest-mean-of-exemplar classifier, preserving old knowledge and stable decision boundaries across tasks.

\textbf{BiC} introduces a lightweight bias-correction layer to compensate for logit shifts from old/new class imbalance, trained on a held-out validation set to reduce bias accumulation.

\textbf{IL2M} addresses score decay in old classes by leveraging stored classifier norms and historical prediction statistics to rescale logits post-hoc, reducing imbalance-induced degradation.

\textbf{DER} adopts a dynamically expandable architecture that freezes previously learned feature extractors and allocates new branches for incoming tasks, preserving prior knowledge by structural isolation rather than distillation alone.
All methods follow their original protocols, with MT-CC applied as a post-hoc calibration module on top of each trained backbone.

\subsection{Details of Calibration Metrics}
\label{sec:calibration_metric}
This section provides detailed definitions of the calibration metrics used
in the main paper. We also introduce the multiclass Brier score, which
evaluates probabilistic predictions without binning, and report the
corresponding results later in Appendix~\ref{sec:brier}.

\noindent\textbf{Expected Calibration Error (ECE)} partitions predictions into $Q$ confidence bins and computes
\[
\mathrm{ECE} = \sum_{i=1}^{Q} \frac{|B_i|}{N} \, | A_i - C_i |,
\]
where $|B_i|$ is the number of samples in bin $i$, $N$ is the total number of samples, and $A_i$ and $C_i$ denote the average accuracy and confidence in bin $i$, respectively.

\noindent\textbf{Adaptive Expected Calibration Error (AECE)} employs adaptive binning based on the cumulative confidence distribution within each test domain:
\[
\mathrm{AECE} = \sum_{i=1}^{Q} \frac{|\tilde{B}_i|}{N} \, | A_i - C_i |,
\]
where $|\tilde{B}_i|$ denotes the number of samples in adaptive bin $i$.

\noindent\textbf{Static Calibration Error (SCE)} extends calibration evaluation to the multiclass setting:
\[
\mathrm{SCE} = \frac{1}{K} \sum_{k=1}^{K} \sum_{b=1}^{Q} \frac{n_{bk}}{N}
\left| \mathrm{acc}(b,k) - \mathrm{conf}(b,k) \right|,
\]
where $K$ is the number of classes, $n_{bk}$ is the number of predictions in bin $b$ for class $k$, and $\mathrm{acc}(b,k)$ and $\mathrm{conf}(b,k)$ are the accuracy and confidence of that bin.

\noindent\textbf{Brier Score} measures the mean squared difference between the predicted probability distribution and the one-hot encoded ground-truth label:
\[
\mathrm{Brier}
=
\frac{1}{N}
\sum_{n=1}^{N}
\sum_{k=1}^{K}
\left(
p_{n,k}-\mathbf{1}[y_n=k]
\right)^2,
\]
where $p_{n,k}$ is the predicted probability of class $k$ for sample $n$, and $\mathbf{1}[y_n=k]$ is one if the ground-truth label $y_n$ is class $k$ and zero otherwise. A lower Brier score indicates better probabilistic predictions.

\clearpage

\section{Additional Ablation Studies}
\vspace{-1mm}
\subsection{Ablation on the Number of Groups}
\label{sec:number_of_groups}
\begin{table}[ht]
    \centering
    \caption{
    Ablation on the number of groups $M$ on CIFAR-100 over five seeds.
    $M=1$ corresponds to single-temperature scaling (T-CIL).
    }
    \label{tab:group_number_ablation}
    \small
    \begin{tabular}{c|ccc}
        \toprule
        $M$ & ECE $\downarrow$ & AECE $\downarrow$ & Brier $\downarrow$ \\
        \midrule
        1 (T-CIL) & $6.03{\pm}0.42$ & $6.04{\pm}0.35$ & $0.570{\pm}0.005$ \\
        \rowcolor{blue!15} 2 & $\mathbf{4.49{\pm}0.53}$ & $\mathbf{4.44{\pm}0.49}$ & $\mathbf{0.556{\pm}0.006}$ \\
        3          & $4.51{\pm}0.47$ & $4.48{\pm}0.45$ & $0.558{\pm}0.007$ \\
        4          & $4.50{\pm}0.41$ & $4.50{\pm}0.42$ & $0.559{\pm}0.007$ \\
        5          & $4.79{\pm}0.40$ & $4.77{\pm}0.41$ & $0.561{\pm}0.007$ \\
        \bottomrule
    \end{tabular}
\end{table}
\noindent We evaluate the sensitivity of MT-CC to the number of groups $M$ by varying $M$ from 1 to 5, where $M=1$ corresponds to single-temperature
scaling (T-CIL). As shown in Table~\ref{tab:group_number_ablation}, nearly all calibration improvement occurs from $M=1$ to $M=2$ (ECE: $6.03 \rightarrow 4.49$); $M=3$ and $M=4$ provide no meaningful additional gains across metrics, while $M=5$ degrades performance. Thus, our choice of $M=2$ is based on the observed saturation of performance rather than evaluating only two groups.

One possible reason is that increasing $M$ introduces more group-specific temperatures to optimize, making the joint routing-temperature optimization more complex. At the same time, the calibration data are partitioned among more groups, leaving fewer samples to reliably estimate each temperature, particularly for sparsely populated groups. Consistent with this explanation, when $M>2$, the learned routing effectively collapses to two dominant groups, leaving the remaining groups sparsely populated and their temperatures poorly estimated.

\vspace{-3mm}
\subsection{Ablation on the Replay Memory Size}

\begin{figure}[ht]
    \centering
    \begin{subfigure}[b]{0.33\linewidth}
        \centering
        \includegraphics[width=\linewidth]{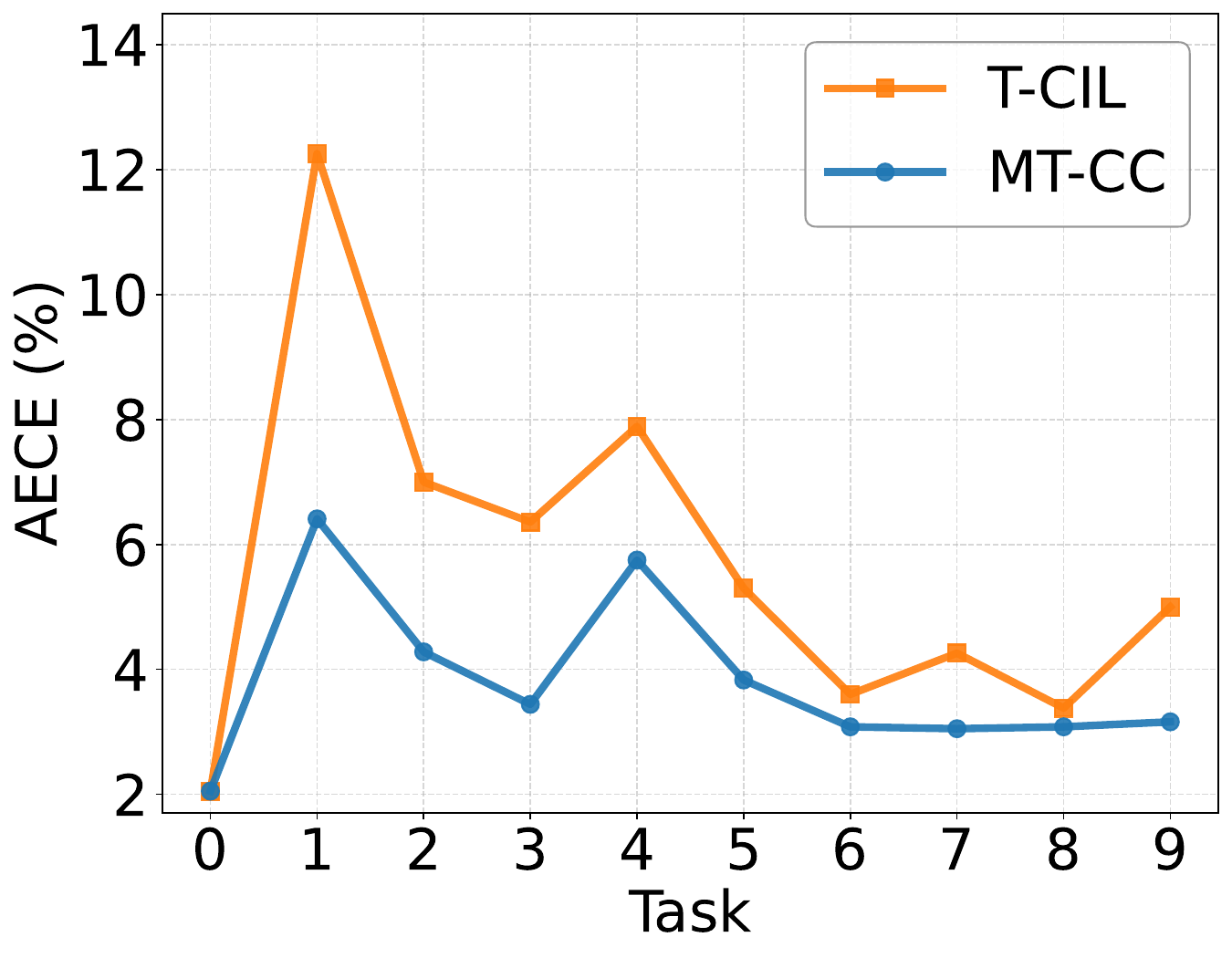}
        \caption{Replay buffer 3K ECE}
        \label{fig:replay_buffer_3000_ece_cifar100}
    \end{subfigure}
    \begin{subfigure}[b]{0.33\linewidth}
        \centering
        \includegraphics[width=\linewidth]{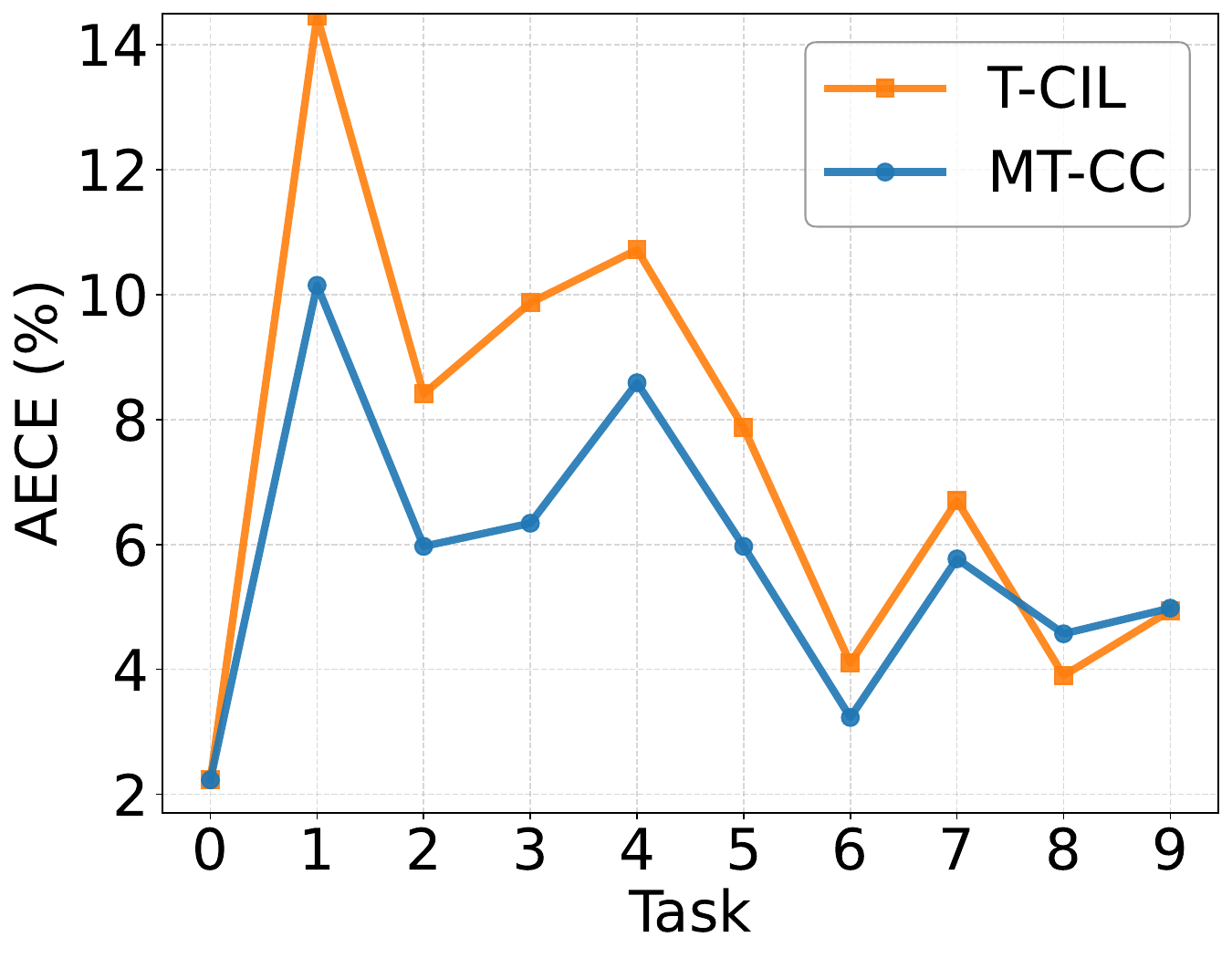}
        \caption{Replay buffer 4K ECE}
        \label{fig:replay_buffer_4000_ece_cifar100}
    \end{subfigure}
    \vspace{-1mm}
    \caption{Ablation study on replay buffer size on CIFAR-100. We report ECE for each task when using a 3K or 4K replay memory.}
    \label{fig:replay_buffer_ablation}
\end{figure}

\noindent We vary the replay buffer size while keeping the same CIFAR-100 experimental setup as in the main paper.
Across all buffer configurations, our method maintains consistently strong calibration performance.
Moreover, MT-CC outperforms T-CIL on nearly all tasks, demonstrating its robustness to memory size changes.

\clearpage

\subsection{Comparison of ECE, AECE, SCE Across Sequential Tasks on CIFAR-100}
\begin{figure}[ht]
    \centering
    \begin{subfigure}[b]{0.32\linewidth}
        \centering
        \includegraphics[width=\linewidth]{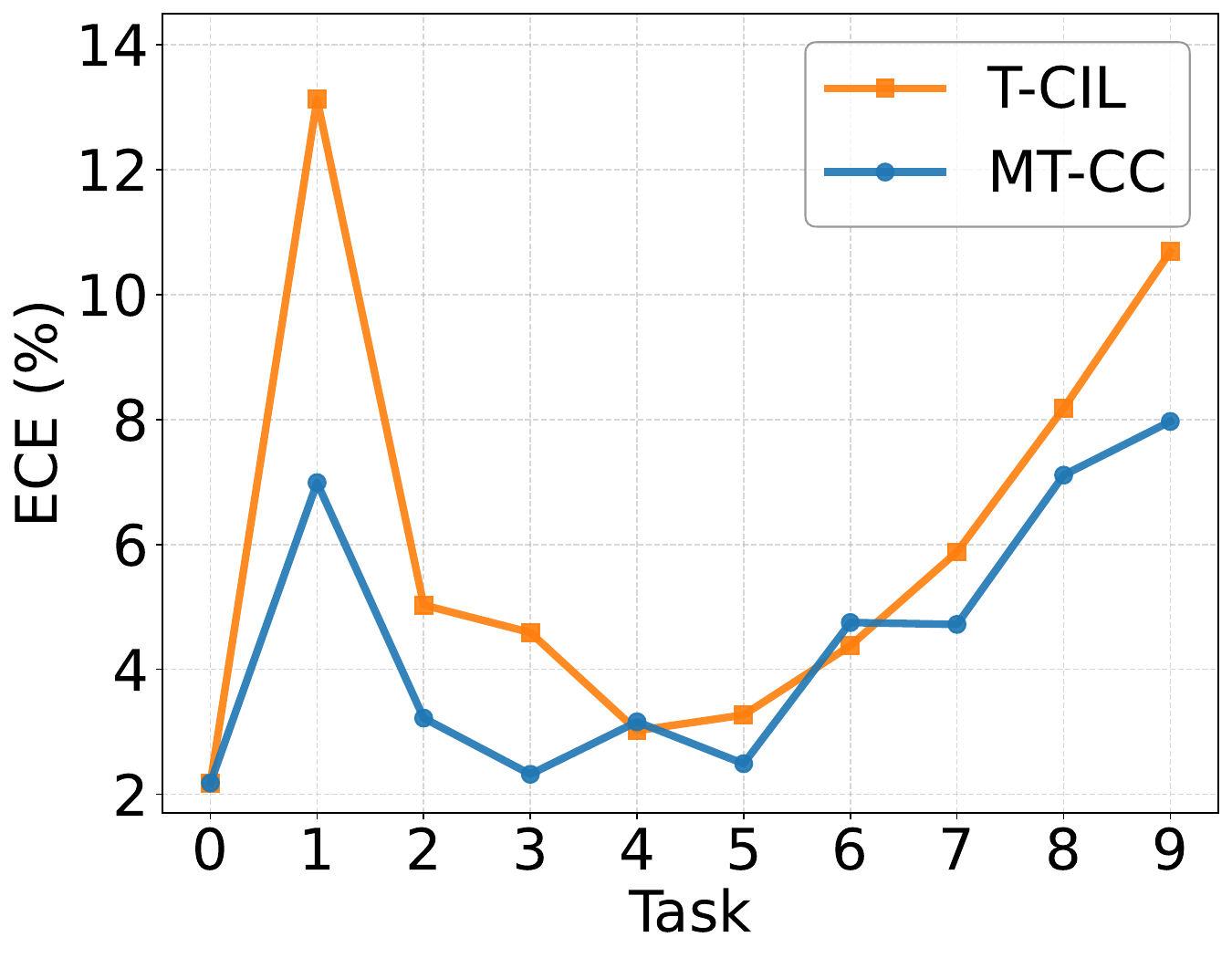}
        \caption{ECE}
        \label{fig:ece_cifar100}
    \end{subfigure}
    \hfill
    \begin{subfigure}[b]{0.32\linewidth}
        \centering
        \includegraphics[width=\linewidth]{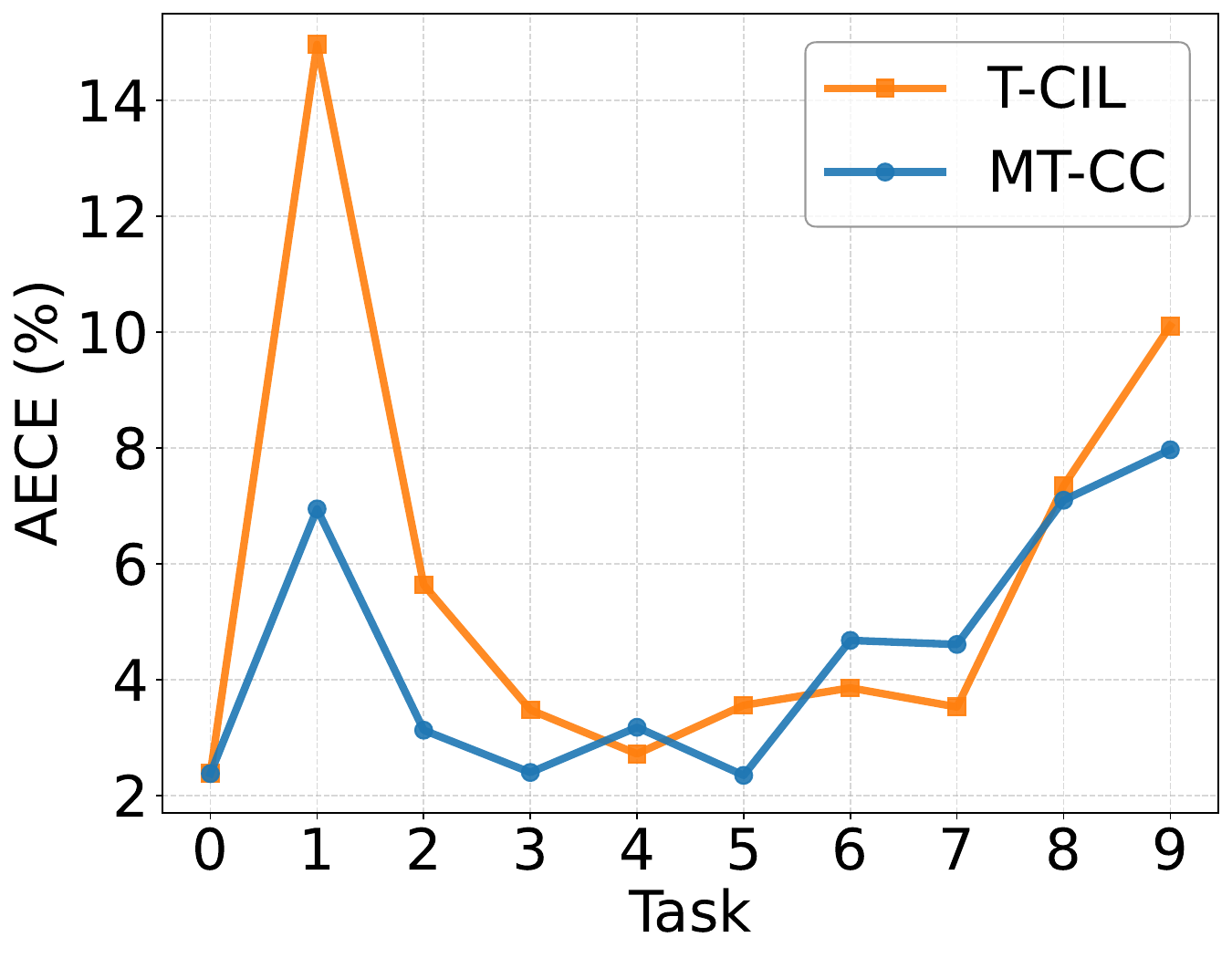}
        \caption{AECE}
        \label{fig:aece_cifar100}
    \end{subfigure}
    \hfill
    \begin{subfigure}[b]{0.32\linewidth}
        \centering
        \includegraphics[width=\linewidth]{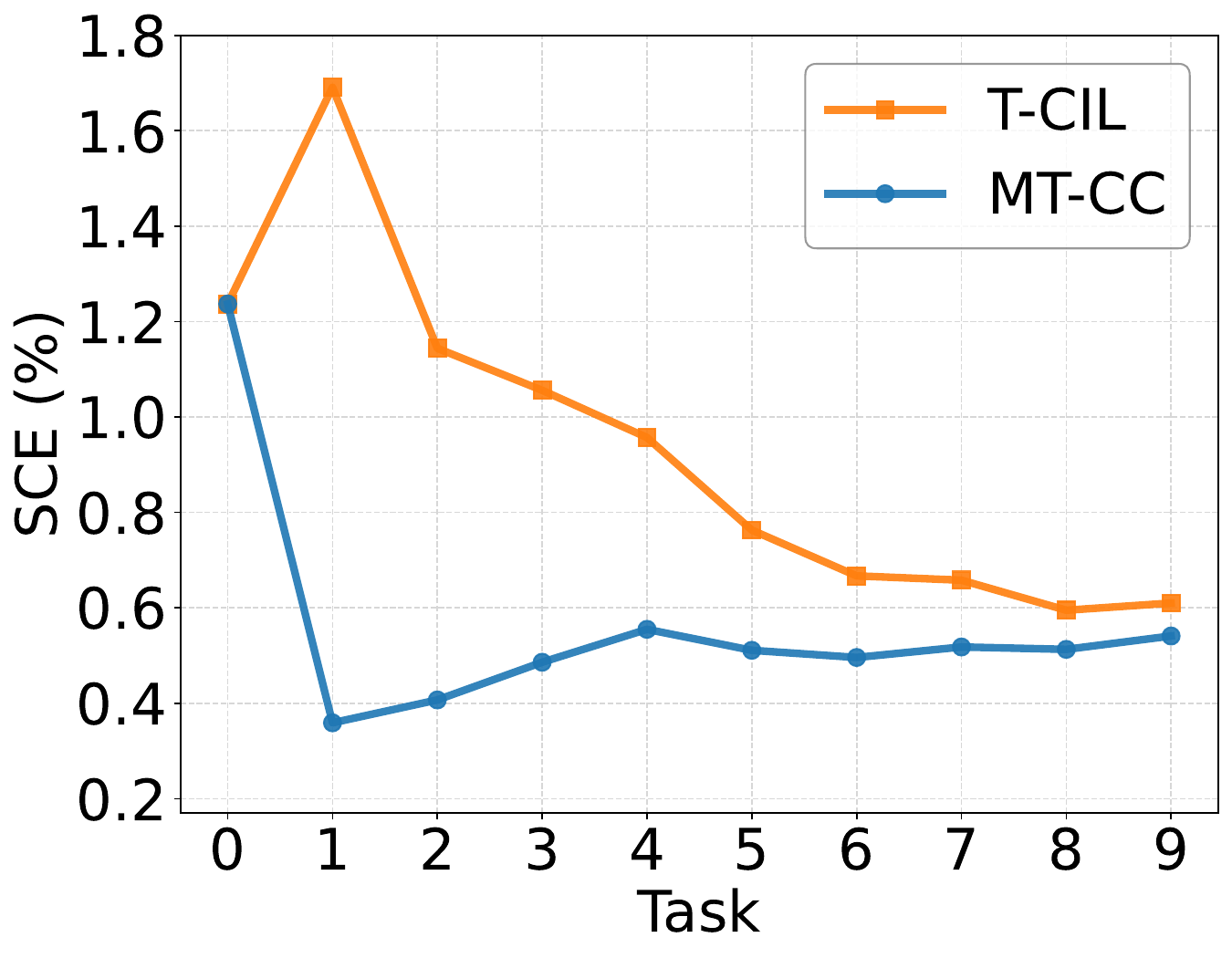}
        \caption{SCE}
        \label{fig:sce_cifar100}
    \end{subfigure}
    \vspace{-2mm}
    \caption{Comparison of calibration metrics across sequential tasks using CIFAR-100.}
    \label{fig:calibration_comparison_cifar100}
    \vspace{-2mm}
\end{figure}

\noindent Figure~\ref{fig:calibration_comparison_cifar100} presents a comparison of ECE, SCE, and AECE across sequential tasks on CIFAR-100. MT-CC consistently achieves lower calibration errors than T-CIL across all three metrics throughout the continual learning process.

\subsection{Results on CIFAR-100: 60-5 Split}
\begin{figure}[ht]
\centering

\begin{minipage}[t]{0.45\linewidth}
    \vspace{0pt}
    \centering
    \includegraphics[width=\linewidth]{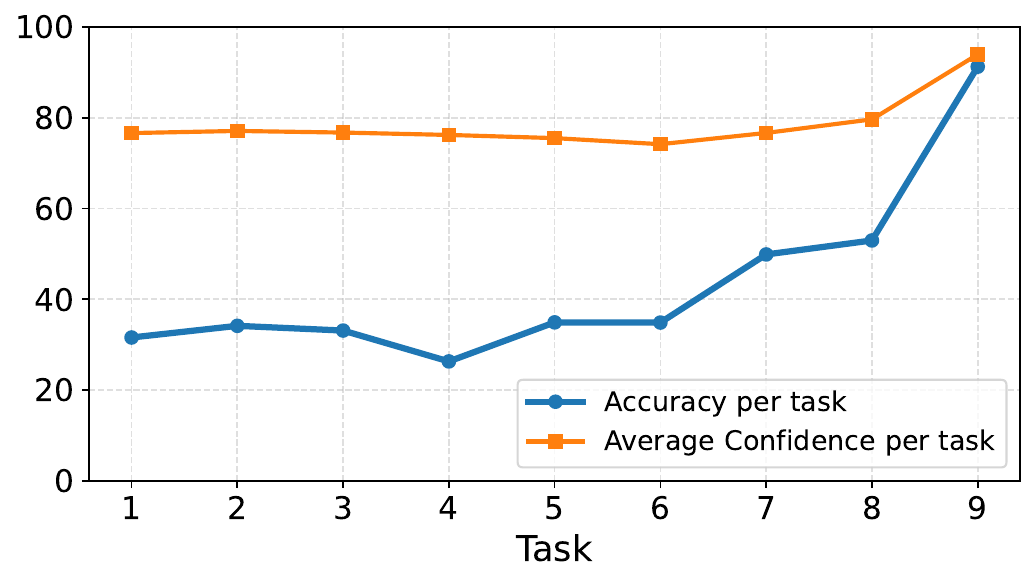}
    \caption{Per-task accuracy and confidence on the final step of the CIFAR-100 60-5 split, illustrating the discrepancy between confidence and accuracy across tasks.}
    \label{fig:cifar100_60_5}
\end{minipage}
\hfill
\begin{minipage}[t]{0.52\linewidth}
    \vspace{0pt}
    \centering
    \begin{adjustbox}{width=\linewidth}
    \begin{tabular}{l|ccc}
    \toprule
    \textbf{Method}
    & \textbf{Acc.\ ($\uparrow$)}
    & \textbf{ECE ($\downarrow$)}
    & \textbf{AECE ($\downarrow$)} \\
    \midrule
    Vanilla     & \meanstdsmall{54.84}{0.41}  & \meanstdsmall{32.45}{0.35}   & \meanstdsmall{32.45}{0.36}   \\
    TS          & \meanstdsmall{54.84}{0.41}  & \meanstdsmall{21.15}{0.84}   & \meanstdsmall{21.12}{0.84}  \\
    ETS         & \meanstdsmall{54.84}{0.41}  & \meanstdsmall{22.32}{0.46}   & \meanstdsmall{22.31}{0.45}   \\
    IRM         & \meanstdsmall{54.84}{0.41}  & \meanstdsmall{23.29}{0.54}   & \meanstdsmall{23.28}{0.53}   \\
    \midrule
    T-CIL       & \meanstdsmall{54.84}{0.41}  & \meanstdsmall{5.48}{0.32}   & \meanstdsmall{5.48}{0.33}   \\
    T-CIL + ETS & \meanstdsmall{54.84}{0.41}  & \meanstdsmall{5.89}{0.34}   & \meanstdsmall{5.87}{0.38}   \\
    T-CIL + IRM & \meanstdsmall{54.84}{0.41}  & \meanstdsmall{6.98}{0.69}   & \meanstdsmall{7.00}{0.66}   \\
    \midrule
    \rowcolor{blue!15}
    \textbf{MT-CC}
    & \meanstdsmall{54.84}{0.41}
    & \boldmath\meanstdsmall{4.79}{0.31}
    & \boldmath\meanstdsmall{4.80}{0.30}  \\
    \bottomrule
    \end{tabular}
    \end{adjustbox}
    \captionof{table}{Calibration performance on CIFAR-100 under the 60-5 split.
    All values are averaged over all nine steps.}
    \label{tab:cifar100_60_5}
\end{minipage}

\end{figure}

We further evaluate our method on the CIFAR-100 60-5 split, where the model is first trained on 60 base classes and then updated over eight incremental steps of five new classes each. This setting enlarges the initial task and induces a more imbalanced class distribution than the standard 10-task protocol, allowing us to verify that the proposed calibration framework remains effective under a different and more challenging CIL configuration. Except for the change in the number of classes per step, experimental settings follow the standard CIFAR-100 protocol.

Figure~\ref{fig:cifar100_60_5} compares per-task accuracy and confidence after the final incremental step.
The results show pronounced discrepancies across tasks, where confidence remains consistently high while accuracy varies substantially, revealing stronger asymmetric miscalibration than in the standard 10-task configuration.
This imbalance increases calibration difficulty because the model assigns similar confidence levels to tasks that differ widely in accuracy.

\noindent Despite this more challenging regime, MT-CC achieves the lowest ECE and AECE among all baselines in Table~\ref{tab:cifar100_60_5}.
MT-CC surpasses T-CIL as well as T-CIL combined with ETS or IRM. This shows that learned grouping and multi-temperature calibration generalize beyond the standard protocol and remain effective under enlarged and imbalanced task structures.

\clearpage
\subsection{Robustness to ECE Binning and Brier Score}
\label{sec:brier}
\vspace{-2mm}
\label{sec:ece_binning_brier}

\begin{table}[ht]
    \centering
    \caption{Calibration performance on CIFAR-10, CIFAR-100, and Tiny-ImageNet
    under different numbers of ECE bins and the multiclass Brier score.
    Lower values indicate better calibration. Results are averaged over five seeds.}
    \label{tab:ece_binning_brier}
    \resizebox{\linewidth}{!}{
    \begin{tabular}{lccc|ccc|ccc}
        \toprule
        & \multicolumn{3}{c|}{\textbf{CIFAR-10}}
        & \multicolumn{3}{c|}{\textbf{CIFAR-100}}
        & \multicolumn{3}{c}{\textbf{Tiny-ImageNet}} \\
        \cmidrule(lr){2-4}
        \cmidrule(lr){5-7}
        \cmidrule(lr){8-10}
        \textbf{Method}
        & ECE@15 $\downarrow$ & ECE@20 $\downarrow$ & Brier $\downarrow$
        & ECE@15 $\downarrow$ & ECE@20 $\downarrow$ & Brier $\downarrow$
        & ECE@15 $\downarrow$ & ECE@20 $\downarrow$ & Brier $\downarrow$ \\
        \midrule

        T-CIL
        & \meanstdsmall{15.36}{1.39}
        & \meanstdsmall{15.37}{1.40}
        & \meanstdsmall{0.498}{0.019}
        & \meanstdsmall{6.19}{0.39}
        & \meanstdsmall{6.27}{0.38}
        & \meanstdsmall{0.570}{0.005}
        & \meanstdsmall{19.04}{0.42}
        & \meanstdsmall{19.13}{0.41}
        & \meanstdsmall{0.883}{0.004} \\

        T-CIL+ETS
        & \meanstdsmall{14.01}{1.31}
        & \meanstdsmall{14.05}{1.30}
        & \meanstdsmall{0.497}{0.019}
        & \meanstdsmall{5.80}{0.45}
        & \meanstdsmall{5.87}{0.44}
        & \meanstdsmall{0.571}{0.004}
        & \meanstdsmall{22.21}{0.21}
        & \meanstdsmall{22.25}{0.21}
        & \meanstdsmall{0.901}{0.003} \\

        T-CIL+IRM
        & \meanstdsmall{14.90}{0.95}
        & \meanstdsmall{14.83}{1.11}
        & \meanstdsmall{0.506}{0.019}
        & \meanstdsmall{6.88}{0.66}
        & \meanstdsmall{6.97}{0.67}
        & \meanstdsmall{0.571}{0.006}
        & \meanstdsmall{18.08}{0.42}
        & \meanstdsmall{18.05}{0.51}
        & \meanstdsmall{0.884}{0.005} \\

        T-CIL+AdaTS
        & \meanstdsmall{14.31}{1.44}
        & \meanstdsmall{14.34}{1.44}
        & \meanstdsmall{0.501}{0.023}
        & \meanstdsmall{7.76}{0.50}
        & \meanstdsmall{7.86}{0.51}
        & \meanstdsmall{0.586}{0.004}
        & \meanstdsmall{21.75}{1.25}
        & \meanstdsmall{21.80}{1.28}
        & \meanstdsmall{0.905}{0.012} \\

        \rowcolor{blue!15}
        \textbf{MT-CC}
        & \boldmath\meanstdsmall{10.22}{1.51}
        & \boldmath\meanstdsmall{10.27}{1.48}
        & \boldmath\meanstdsmall{0.447}{0.015}
        & \boldmath\meanstdsmall{4.62}{0.51}
        & \boldmath\meanstdsmall{4.72}{0.54}
        & \boldmath\meanstdsmall{0.556}{0.006}
        & \boldmath\meanstdsmall{15.87}{0.37}
        & \boldmath\meanstdsmall{15.93}{0.36}
        & \boldmath\meanstdsmall{0.839}{0.003} \\

        \bottomrule
    \end{tabular}
    }
\end{table}

To ensure that the observed calibration gains are not artifacts of specific metric configurations, we investigate the binning sensitivity of our evaluation. Specifically, we re-evaluate all methods on CIFAR-100 using 15 and 20 bins for ECE. Furthermore, to provide a strictly binning-free assessment, we incorporate the multiclass Brier score. We use the same trained models and calibrated probabilities across all evaluations, and report the mean and standard deviation over five seeds.

As shown in Table~\ref{tab:ece_binning_brier}, MT-CC consistently achieves the lowest ECE across all three bin configurations, and the relative ranking of the methods remains unchanged as the number of bins increases. Crucially, MT-CC also obtains the lowest multiclass Brier score, confirming its overall robustness.

\vspace{-2mm}
\subsection{Applicability to Static Classification}
\vspace{-2mm}
\label{sec:static_classification}

\begin{table}[ht]
    \centering
    \caption{Comparison of single-temperature scaling and MT-CC in static and class-incremental learning settings on CIFAR-100 over five seeds.}
    \label{tab:static_classification}
    \resizebox{0.5\linewidth}{!}{
        \begin{tabular}{lcc}
            \toprule
            Setting & Single-T ECE $\downarrow$ & MT-CC ECE $\downarrow$ \\
            \midrule
            Static joint training
                & $0.87 \pm 0.14$
                & $0.87 \pm 0.14$ \\
            CIL (task average)
                & $6.03 \pm 0.42$
                & $\mathbf{4.49 \pm 0.53}$ \\
            \bottomrule
        \end{tabular}
    }
\end{table}

Although groupwise temperature scaling is applicable to static classification, its benefit in MT-CC arises from modeling the asymmetric calibration behavior induced by continual learning. To verify that the improvement does not simply result from using multiple temperatures, we apply the same calibration pipeline to a jointly trained static CIFAR-100 model.

As shown in Table~\ref{tab:static_classification}, MT-CC provides no additional improvement over single-temperature scaling in the static setting. The learned router effectively collapses to a single group. Specifically, the fraction of samples assigned to the smaller group is $0.000 \pm 0.000$, and the active-group temperature differs from the single-temperature solution by less than 0.003 in every seed. MT-CC therefore reduces to standard single-temperature scaling in this setting.

In contrast, under CIL, MT-CC reduces the task-averaged ECE from $6.03$ to $4.49$. This comparison confirms that MT-CC effectively targets the unique calibration shifts inherent to CIL, rather than merely benefiting from increased parameter capacity.

\vspace{-1mm}
\subsection{Effect of Validation Set Size}

\begin{table}[ht]
\vspace{-3mm}
\centering
\caption{Calibration performance under varying validation set sizes (lower is better).}
\label{tab:val_size}
\renewcommand{\arraystretch}{1.15}
\begin{adjustbox}{max width=0.7\linewidth}
\begin{tabular}{c|ccccc}
\toprule
{\small\textbf{Validation Size}}
& {\small\textbf{0}}
& {\small\textbf{500}}
& {\small\textbf{1000}}
& {\small\textbf{1500}}
& {\small\textbf{2000}} \\
\midrule
T-CIL
& \meanstdsmall{16.13}{0.19}
& \meanstdsmall{6.19}{0.66}
& \meanstdsmall{6.38}{0.47}
& \meanstdsmall{6.24}{0.39}
& \meanstdsmall{6.03}{0.42} \\
\rowcolor{blue!15}\textbf{MT-CC}
& \boldmath\meanstdsmall{15.96}{1.05}
& \boldmath\meanstdsmall{5.92}{0.95}
& \boldmath\meanstdsmall{6.04}{0.78}
& \boldmath\meanstdsmall{5.89}{0.77}
& \boldmath\meanstdsmall{4.49}{0.53} \\
\bottomrule
\end{tabular}
\end{adjustbox}
\end{table}

\noindent Table~\ref{tab:val_size} reports the effect of validation set size.
MT-CC consistently outperforms T-CIL, a state-of-the-art method tailored to temperature
scaling in class-incremental learning.
Even under limited validation data regimes, MT-CC consistently outperforms T-CIL, demonstrating robustness to calibration data scarcity.
Notably, even when the validation set size is 0, MT-CC achieves an ECE of 15.96 compared
to 16.13 for T-CIL, relying solely on samples from the current task for grouping network training and temperature scaling.

\clearpage

\subsection{Inference Time Analysis}
\vspace{-3mm}
\begin{table}[ht]
\centering
\setlength{\tabcolsep}{10pt}
\caption{The average per-sample inference time across all tasks on ImageNet over 3 seeds. Relative time is reported with TS as the baseline (1$\times$). MT-CC requires negligible additional inference overhead compared to standard TS.}

\label{tab:inference_time}
\resizebox{0.5\linewidth}{!}{
\begin{tabular}{lcc}
\toprule
Method & Per-sample inference time ($\mu$s) & Relative \\
\midrule
TS & 1250.41 & 1.0000$\times$ \\
PTS & 1250.65 & 1.0002$\times$ \\
AdaTS & 1252.63 & 1.0018$\times$ \\
\rowcolor{blue!15} MT-CC & 1250.80 & 1.0003$\times$ \\
\bottomrule
\end{tabular}
}
\end{table}

\noindent{Table~\ref{tab:inference_time} reports the average inference time across all tasks on ImageNet in microsecond scale, measured on a single NVIDIA RTX A6000 GPU.
AdaTS incurs slightly higher inference cost than others due to the use of a class-conditional VAE.
In contrast, MT-CC introduces negligible additional overhead compared to standard temperature scaling (TS).
Since the grouping module is a lightweight network, the additional latency introduced by MT-CC is negligible in practice.}

\subsection{Comparison with Feature-Matched and Grouped Calibration Methods}
\label{sec:feature_matched_comparison}

\noindent MT-CC outperforms both feature-matched instance-wise temperature scaling and unsupervised group-wise temperature scaling, showing that its improvement cannot be attributed solely to the CIL-aware inputs or the use of multiple temperatures. We evaluate progressively stronger baselines that isolate the contributions of the input features, heterogeneous calibration groups, and learned calibration-aware routing.

\begin{table}[ht]
    \centering
    \caption{Comparison with feature-matched and grouped calibration methods on CIFAR-100 over 5 seeds. The CIL-aware setting uses the same feature vector as MT-CC.}
    \label{tab:feature_matched_comparison}
    \resizebox{\linewidth}{!}{
    \begin{tabular}{lllccc}
        \toprule
        \textbf{Method}
        & \textbf{Calibration / Routing}
        & \textbf{Feature}
        & \textbf{ECE $\downarrow$}
        & \textbf{AECE $\downarrow$}
        & \textbf{Brier $\downarrow$} \\
        \midrule

        PTS
        & Instance-wise TS
        & Top-10 logits
        & $9.87 \pm 0.77$
        & $10.14 \pm 0.67$
        & $0.596 \pm 0.008$ \\

        PTS
        & Instance-wise TS
        & CIL-aware features
        & $6.88 \pm 0.91$
        & $6.87 \pm 0.91$
        & $0.575 \pm 0.008$ \\

        \midrule

        AdaTS
        & Instance-wise TS
        & Penultimate feature
        & $7.67 \pm 0.52$
        & $7.66 \pm 0.49$
        & $0.586 \pm 0.004$ \\

        AdaTS
        & Instance-wise TS
        & CIL-aware features
        & $6.55 \pm 0.74$
        & $6.52 \pm 0.75$
        & $0.574 \pm 0.004$ \\

        \midrule

        GC+TS \citep{yang2023beyond}
        & Semantic-aware learned grouping + group-wise TS
        & Penultimate feature
        & $7.23 \pm 0.25$
        & $7.24 \pm 0.26$
        & $0.576 \pm 0.005$ \\

        K-means Group-TS
        & K-means ($K=2$) + group-wise TS
        & CIL-aware features
        & $5.41 \pm 0.51$
        & $5.37 \pm 0.51$
        & $0.561 \pm 0.005$ \\

        \rowcolor{blue!15}\textbf{MT-CC}
        & Learned routing + group-wise TS
        & CIL-aware features
        & $\mathbf{4.49 \pm 0.53}$
        & $\mathbf{4.44 \pm 0.49}$
        & $\mathbf{0.556 \pm 0.006}$ \\

        \bottomrule
    \end{tabular}
    }
\end{table}

\noindent\textbf{Contribution of CIL-aware features.}
CIL-aware features provide more informative calibration signals under continual distribution shifts. We provide PTS and AdaTS with the same input vector used by MT-CC, $v(x)=[\delta,p_{\mathrm{old}},p_{\mathrm{new}},p]$. Both methods improve over their original feature settings, confirming the effectiveness of the proposed input design.

\noindent\textbf{Contribution of asymmetric calibration groups.}
Explicitly modeling asymmetric calibration groups improves upon instance-wise temperature prediction. K-means Group-TS clusters the same CIL-aware feature vectors into two groups, optimizes a separate temperature for each cluster, and assigns each test sample the temperature of its nearest centroid. It outperforms the feature-matched PTS and AdaTS variants, supporting the use of group-specific calibration.

\noindent\textbf{Contribution of learned calibration-aware routing.}
Learning groups according to the calibration objective is more effective than clustering the feature space independently of calibration. MT-CC outperforms K-means Group-TS even though both use the same input features and two group-specific temperatures.

\noindent\textbf{Comparison with existing grouping calibration.}
We further compare MT-CC with GC+TS~\citep{yang2023beyond}, a prior
semantic-aware grouping calibration method developed for static classification.
GC+TS learns a grouping function from deep semantic features and applies a
separate temperature to each learned group. The comparison shows that CIL-aware grouping is more effective for capturing calibration heterogeneity under continual task shifts.

\subsection{CIL Changes the Calibration Meaning of Logit Margin}
\label{sec:margin_temperature}

\noindent The logit margin is strongly associated with the required calibration correction under CIL, whereas this relationship is nearly absent in static classification. For a jointly trained static CIFAR-100 model and the CIL model at the final task, we divide samples into 10 equal-count margin bins and independently fit the NLL-optimal temperature $T^*$ in each bin.

\begin{table}[ht]
    \centering
    \caption{NLL-optimal temperature $T^*$ across logit-margin bins on CIFAR-100 averaged over 5 seeds.}
    \label{tab:margin_temperature}
    \resizebox{0.8\linewidth}{!}{
    \begin{tabular}{lcccccccccc}
        \toprule
        \textbf{Setting}
        & \textbf{b0}
        & \textbf{b1}
        & \textbf{b2}
        & \textbf{b3}
        & \textbf{b4}
        & \textbf{b5}
        & \textbf{b6}
        & \textbf{b7}
        & \textbf{b8}
        & \textbf{b9} \\
        & \textbf{(low)}
        & & & & & & & & 
        & \textbf{(high)} \\
        \midrule
        Static
        & 1.98 & 1.99 & 2.00 & 1.95 & 2.00
        & 1.94 & 1.96 & 1.92 & 2.04 & 2.10 \\
        CIL
        & 1.58 & 1.68 & 1.77 & 1.84 & 2.00
        & 2.13 & 2.36 & 2.59 & 2.99 & 3.82 \\
        \bottomrule
    \end{tabular}
    }
\end{table}

In the static model, $T^*$ remains nearly constant across margin bins even though accuracy increases with the margin. Under CIL, however, catastrophic forgetting degrades accuracy without a corresponding drop in prediction margin, so high-margin predictions become increasingly overconfident. Thus, the logit margin becomes a CIL-aware signal of the calibration correction required by different samples.

\subsection{Complementarity of Margin and Confidence}
\label{sec:margin_confidence_complementarity}

\noindent Confidence provides additional calibration information even after the logit margin is controlled. We divide samples into two equal-count margin regions and further split each region into low- and high-confidence halves. We independently fit $T^*$ to each subset and define $\Delta T^*$ as the difference between the optimal temperatures of the high- and low-confidence subsets.

\begin{table}[ht]
    \centering
    \caption{Optimal temperatures after controlling for the logit margin.}
    \label{tab:margin_confidence_temperature}
    \begin{tabular}{llccc}
        \toprule
        \textbf{Setting}
        & \textbf{Margin region}
        & \textbf{Low-conf. $T^*$}
        & \textbf{High-conf. $T^*$}
        & \boldmath$\Delta T^*$ \\
        \midrule
        Static
        & Low
        & 1.98 & 1.99 & 0.01 \\
        & High
        & 1.94 & 2.03 & 0.09 \\
        \midrule
        CIL
        & Low
        & 1.60 & 1.95 & 0.34 \\
        & High
        & 2.26 & 3.32 & 1.05 \\
        \bottomrule
    \end{tabular}
\end{table}

Confidence produces only small temperature differences in the static model but substantially larger differences under CIL. This result shows that margin and confidence capture complementary aspects of CIL-induced calibration heterogeneity.

% \subsection{CIL-Specificity of Old/New Probability Masses}
% \label{sec:old_new_mass}

% \noindent The old/new probability masses capture calibration-relevant CIL structure beyond generic class-mass statistics. We replace them with probability masses computed from a random partition of the classes while matching the two subset sizes to the true old/new split.

% \begin{table}[ht]
%     \centering
%     \caption{Comparison of true old/new probability masses with a size-matched random class-mass pair on CIFAR-100.}
%     \label{tab:old_new_random_mass}
%     \begin{tabular}{lccc}
%         \toprule
%         \textbf{Grouping-network input}
%         & \textbf{ECE $\downarrow$}
%         & \textbf{AECE $\downarrow$}
%         & \textbf{Brier $\downarrow$} \\
%         \midrule
%         Random class-mass pair
%         & 4.60 & 4.61 & 0.571 \\
%         True old/new pair (MT-CC)
%         & \textbf{4.49} & \textbf{4.44} & \textbf{0.556} \\
%         \bottomrule
%     \end{tabular}
% \end{table}

% The true old/new pair improves all three calibration metrics over the size-matched random partition. This confirms that its benefit arises from CIL-specific old/new structure rather than merely from adding two class-mass statistics.
\clearpage

\subsection{Benefit of Explicit Old/New Class Structure}
\label{sec:explicit_old_new}

\begin{table}[t]
\centering
\caption{
Benefit of explicitly providing the old/new class structure on CIFAR-100 across different grouping-set sizes.
The optimization budget is matched across data sizes and $S_{\mathrm{cal}}$ is fixed.
Results report ECE (\%) and Brier score over five seeds.
}
\label{tab:explicit_old_new}

\resizebox{\linewidth}{!}{
\begin{tabular}{c|cc|cc|cc|cc}
\toprule
\multirow{2}{*}{$|S_{\mathrm{grp}}|$}
& \multicolumn{2}{c|}{$[\delta,p_{\mathrm{old}},p_{\mathrm{new}},p]$}
& \multicolumn{2}{c|}{$[\delta,p]$}
& \multicolumn{2}{c|}{$[\delta,p_{\mathrm{rand1}},p_{\mathrm{rand2}},p]$}
& \multicolumn{2}{c}{$[\delta,p_{\mathrm{old}},p_{\mathrm{new}}]$} \\
& ECE & Brier & ECE & Brier & ECE & Brier & ECE & Brier \\
\midrule
50
& {\boldmath\meanstdsmall{4.77}{0.43}}
& {\boldmath\meanstdsmall{0.565}{0.009}}
& \meanstdsmall{4.92}{0.50}
& \meanstdsmall{0.573}{0.006}
& \meanstdsmall{5.00}{0.43}
& \meanstdsmall{0.573}{0.006}
& \meanstdsmall{5.98}{0.48}
& \meanstdsmall{0.572}{0.004} \\

100
& {\boldmath\meanstdsmall{4.36}{0.46}}
& {\boldmath\meanstdsmall{0.560}{0.007}}
& \meanstdsmall{4.69}{0.40}
& \meanstdsmall{0.571}{0.005}
& \meanstdsmall{4.81}{0.45}
& \meanstdsmall{0.571}{0.006}
& \meanstdsmall{5.78}{0.30}
& \meanstdsmall{0.571}{0.005} \\

250
& {\boldmath\meanstdsmall{4.49}{0.45}}
& {\boldmath\meanstdsmall{0.558}{0.006}}
& \meanstdsmall{4.70}{0.27}
& \meanstdsmall{0.572}{0.005}
& \meanstdsmall{4.73}{0.34}
& \meanstdsmall{0.572}{0.006}
& \meanstdsmall{5.74}{0.33}
& \meanstdsmall{0.570}{0.005} \\

500
& {\boldmath\meanstdsmall{4.54}{0.56}}
& {\boldmath\meanstdsmall{0.557}{0.006}}
& \meanstdsmall{4.72}{0.41}
& \meanstdsmall{0.571}{0.006}
& \meanstdsmall{4.71}{0.37}
& \meanstdsmall{0.571}{0.006}
& \meanstdsmall{5.83}{0.30}
& \meanstdsmall{0.571}{0.005} \\

1000
& {\boldmath\meanstdsmall{4.49}{0.53}}
& {\boldmath\meanstdsmall{0.556}{0.006}}
& \meanstdsmall{4.63}{0.30}
& \meanstdsmall{0.571}{0.007}
& \meanstdsmall{4.59}{0.40}
& \meanstdsmall{0.571}{0.007}
& \meanstdsmall{5.82}{0.31}
& \meanstdsmall{0.571}{0.005} \\
\bottomrule
\end{tabular}
}

\end{table}

Although $p_{\mathrm{old}}$ and $p_{\mathrm{new}}$ are linear functions of the
full softmax vector $p$ and therefore do not increase the expressiveness of the
linear grouping network, explicitly providing them exposes the old-versus-new
class structure directly to the router.
We investigate whether explicitly exposing this structure benefits calibration
despite the same information already being contained in $p$.

\textbf{Input variants.}
To isolate the effect of explicitly providing the old/new class structure, we compare the following input configurations:
\begin{itemize}[label=\textbullet, labelindent=0pt, leftmargin=*]

\item \textbf{$[\delta,p_{\mathrm{old}},p_{\mathrm{new}},p]$ :}
The full MT-CC input, which combines the logit margin, explicit old/new probability masses, and the full softmax vector.

\item \textbf{$[\delta,p]$ :}
Removes the explicit old/new probability masses while retaining the logit margin and full softmax vector.

\item \textbf{$[\delta,p_{\mathrm{rand1}},p_{\mathrm{rand2}},p]$ :}
Replaces the old/new probability masses with probability masses computed from a size-matched random partition of the classes.
Since these random class-mass features are also linear functions of $p$, this variant controls for the effect of simply adding redundant aggregate features.

\item \textbf{$[\delta,p_{\mathrm{old}},p_{\mathrm{new}}]$ :}
Retains the logit margin and explicit old/new probability masses while removing the full softmax vector.

\end{itemize}

\textbf{Experimental setup.}
To verify that the benefit of the explicit old/new structure is not specific to
a particular grouping-data regime, we vary the number of distinct samples used
to train the grouping network,
$|S_{\mathrm{grp}}| \in \{50,100,250,500,1000\}$,
while keeping the temperature-fitting subset $S_{\mathrm{cal}}$ fixed.
The same grouping samples are used across input variants at each data size.
We match the optimization budget across grouping-set sizes so that each setting
receives approximately the same number of gradient updates, isolating the
effect of the number of distinct grouping samples.
All other hyperparameters are kept identical.

\textbf{Benefit of explicit old/new structure.}
As shown in Table~\ref{tab:explicit_old_new}, explicitly providing the old/new class structure consistently improves both ECE and Brier score over using only $[\delta,p]$ across grouping-data budgets.
This indicates that the grouping network learns more effective calibration routing when the CIL-specific old/new structure is directly exposed.

\textbf{CIL-specificity of the improvement.}
Replacing the true old/new probability masses with size-matched random class-mass
summaries does not reproduce the same improvement and performs similarly to
$[\delta,p]$.
Thus, the gain cannot be explained merely by adding redundant linear aggregate features. Instead, it depends on explicitly exposing the old-versus-new structure aligned with continual learning.

\textbf{Complementarity with the full softmax vector.}
Using only $[\delta,p_{\mathrm{old}},p_{\mathrm{new}}]$ yields substantially
higher ECE, showing that the old/new summaries do not replace the fine-grained
class-wise information in $p$.
Instead, they provide complementary CIL-aware structure.

\textbf{Sample efficiency.}
The full MT-CC input achieves an ECE of 4.36\% with only 100 grouping samples,
comparable to 4.49\% with 1,000 samples.
This indicates that effective routing can be learned from a relatively small
number of distinct grouping samples when the optimization budget is controlled.

\clearpage
\subsection{Continual Update and Stability of the Grouping Network}
\label{sec:router_update_stability}
\noindent We continually update the grouping network so that it can adapt to the evolving classifier across tasks. Continually updating the grouping network improves calibration while preserving the routing structure learned across tasks. We validate this design choice by comparing alternative update strategies and measuring group-assignment consistency across adjacent tasks.

\begin{table}[ht]
    \centering
    \caption{Average calibration performance of different grouping-network update strategies on CIFAR-100 over five seeds.}
    \label{tab:router_update_strategy}
    \resizebox{0.9\linewidth}{!}{
    \begin{tabular}{lcccc}
        \toprule
        Grouping-network strategy
            & ECE $\downarrow$
            & AECE $\downarrow$
            & SCE ($\times 10^{-3}$) $\downarrow$
            & Brier $\downarrow$ \\
        \midrule
        \rowcolor{blue!15}\textbf{MT-CC}
            & $\mathbf{4.49 \pm 0.53}$
            & $\mathbf{4.44 \pm 0.49}$
            & $\mathbf{5.12 \pm 0.11}$
            & $\mathbf{0.556 \pm 0.006}$ \\
        Lag-1
            & $4.82 \pm 0.36$
            & $4.82 \pm 0.36$
            & $5.30 \pm 0.12$
            & $0.558 \pm 0.006$ \\
        Fresh each task
            & $4.98 \pm 0.35$
            & $4.96 \pm 0.34$
            & $5.38 \pm 0.14$
            & $0.561 \pm 0.007$ \\
        \bottomrule
    \end{tabular}
    }
\end{table}

\textbf{Effectiveness of continual router updates.}
\noindent Updating the previous router is more effective than either freezing it or discarding its previously learned routing structure.
We compare the following three grouping-network update strategies on CIFAR-100:

\begin{itemize}[label=\textbullet, labelindent=0pt, leftmargin=*]
\item \textbf{MT-CC :}
The router learned at task $t-1$ is carried forward and updated using the calibration data available at task $t$.

\item \textbf{Lag-1 :}
The router learned at task $t-1$ is used without further updates to calibrate the model at task $t$, while only the group temperatures are re-fitted.

\item \textbf{Fresh each task :}
The router is reinitialized and trained independently at every task.
\end{itemize}

As shown in Table~\ref{tab:router_update_strategy}, continually updating the router outperforms both freezing and reinitializing it across all calibration metrics. This validates the MT-CC design of retaining previously learned routing information while adapting it to the evolving classifier.

\textbf{Stability of continual router updates.}
Continually updating the router causes little routing drift. When the classifier outputs are held fixed, both groups retain more than 90\% of their assignments across adjacent tasks. We measure group retention over all adjacent task transitions and five seeds. Groups are ordered by their learned temperatures, with G1 denoting the lower-temperature group and G2 the higher-temperature group. For each group, retention is defined as the fraction of samples assigned to that group by the task $t-1$ router that remain in the same group under the task $t$ router.

We evaluate retention under two conditions. In the actual transition, both the classifier and router change from task $t-1$ to task $t$. In the router-only condition, the task $t$ classifier outputs are held fixed, and the task $t-1$ and task $t$ routers are evaluated using the same outputs.

\begin{table}[ht]
    \centering
    \caption{Group retention across adjacent tasks on CIFAR-100, averaged over five seeds. G1 and G2 denote the lower- and higher-temperature groups, respectively.}
    \label{tab:router_consistency}
    \begin{tabular}{lcc}
        \toprule
        Condition
            & G1 retention (\%) $\uparrow$
            & G2 retention (\%) $\uparrow$ \\
        \midrule
        Classifier and router change
            & $71.8 \pm 7.5$
            & $42.0 \pm 12.4$ \\
        Router change only
            & $94.3 \pm 6.3$
            & $90.3 \pm 9.6$ \\
        \bottomrule
    \end{tabular}
\end{table}

As shown in Table~\ref{tab:router_consistency}, the task $t-1$ and task $t$ routers produce highly consistent assignments when evaluated on the same classifier outputs. The lower G2 retention in the actual transition is expected because the introduction of new classes substantially changes the composition of the high-temperature group.

\clearpage

\section{How GDCA Stabilizes Routing-Temperature Optimization}
\label{sec:gdca_stabilization}
\vspace{-3mm}
We investigate how GDCA stabilizes the coupled routing--temperature optimization. By complementing the sample-level NLL with groupwise confidence--accuracy alignment, GDCA helps the joint optimizer distinguish routing solutions with similar likelihood, guides the estimation of group-specific temperatures, and mitigates routing collapse into a single dominant group. Together, these effects explain the calibration improvement reported in the main paper. We first show that NLL alone is insufficient for calibration-relevant routing, then explain the optimization mechanism of GDCA, and finally analyze routing degeneracy with and without GDCA.

\begin{figure}[ht]
    \centering
    \includegraphics[width=0.55\linewidth]{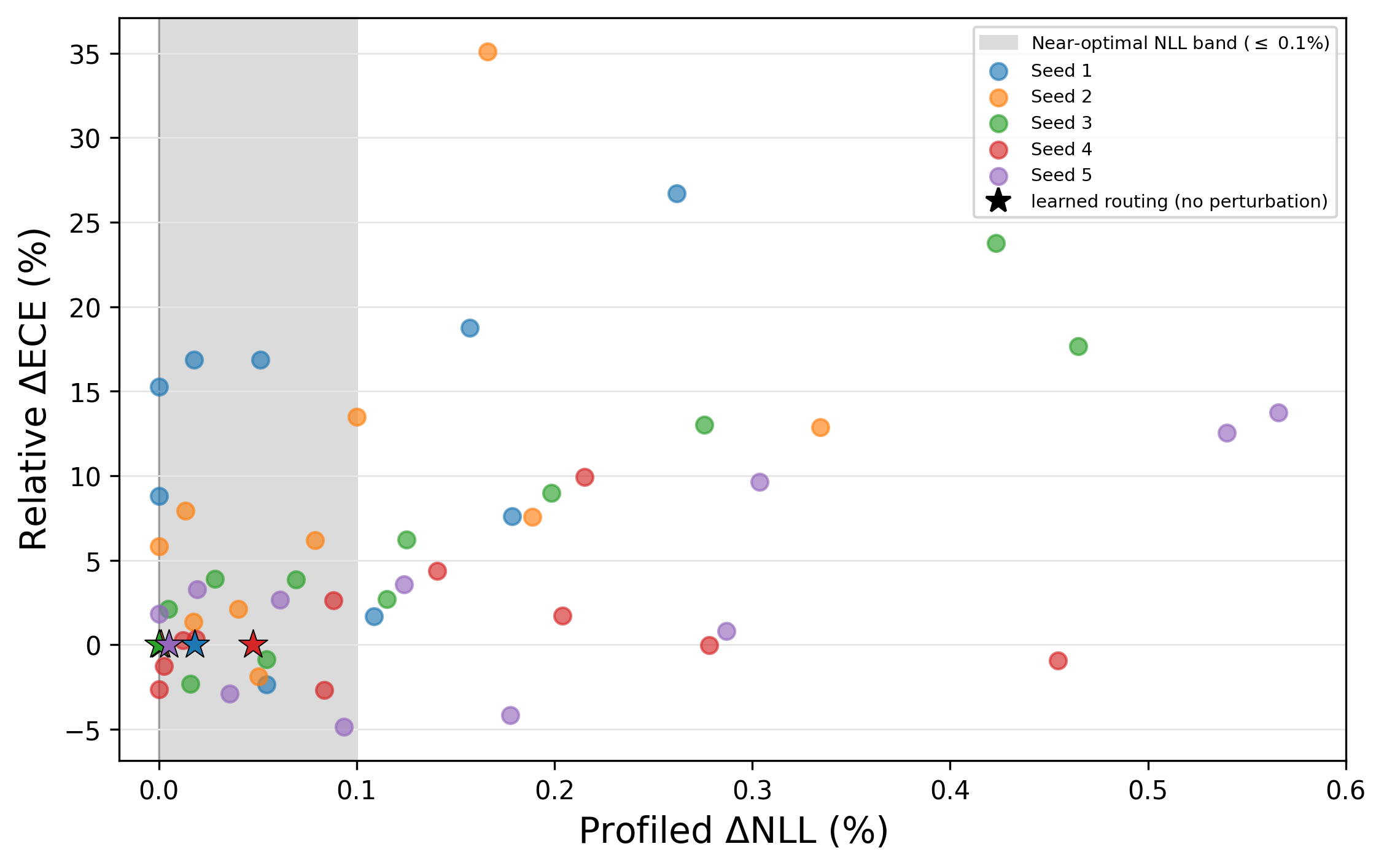}
    \vspace{-3mm}
    \caption{Profile-likelihood analysis of perturbed routing solutions at the final task on CIFAR-100 over five seeds. The group temperatures are re-optimized from scratch for each routing perturbation. For each seed, profiled $\Delta\mathrm{NLL}$ is measured relative to the minimum profiled NLL across all perturbations, while relative $\Delta\mathrm{ECE}$ is measured with respect to the learned routing. The shaded region indicates routing solutions whose profiled NLL lies within $0.1\%$ of the seed-wise minimum.}
    \label{fig:gdca_profile_likelihood}
\end{figure}
\vspace{-3mm}

\textbf{NLL alone is insufficient for calibration-relevant routing.}
Optimizing the routing with NLL alone may not distinguish routing solutions with different calibration quality. Even after re-optimizing the temperatures for each routing configuration, substantially different routing solutions can achieve nearly identical NLL values. To examine this limitation, we perturb the learned routing and re-optimize the temperatures from scratch at each perturbation point.

As shown in Figure~\ref{fig:gdca_profile_likelihood}, some routing solutions within the near-optimal NLL band produce a relative increase of more than $15\%$ in test ECE compared with the learned routing. Thus, NLL can be nearly indifferent among routing solutions with materially different calibration quality, making it insufficient as the sole objective for learning calibration-relevant routing.

\textbf{Optimization mechanism.}
Each group $G_m$ is constructed using a hard routing assignment, and thus gradients do not pass through the group membership or the empirical accuracy $\operatorname{Acc}(G_m)$. Instead, they are propagated through $\operatorname{Conf}(G_m)$, which is computed from the calibrated mixture confidence using the soft routing weights. Consequently, GDCA directly adjusts the group temperatures toward groupwise confidence--accuracy alignment. Once the temperatures become differentiated, the mixture confidence also becomes sensitive to the routing weights, allowing GDCA to update the router indirectly through the mixture confidence while jointly optimizing the temperatures.

\begin{table}[ht]
    \centering
    \caption{Routing-group sizes with and without GDCA over five seeds. The minority-group fraction denotes the proportion of test samples assigned to the less populated of the two groups.}
    \label{tab:gdca_routing_behavior}
    \resizebox{0.7\linewidth}{!}{
    \begin{tabular}{lcc}
        \toprule
        \textbf{Routing behavior}
        & \textbf{w/o GDCA}
        & \textbf{with GDCA} \\
        \midrule
        Frequency of minority-group fraction below $20\%$
        & $20.0\%$
        & $4.4\%$ \\
        Mean minority-group fraction
        & $29.7\%$
        & $37.5\%$ \\
        \bottomrule
    \end{tabular}
    }
\end{table}

\textbf{Routing degeneracy with and without GDCA.}
During training, the router may assign most samples to one dominant group, causing the other group to become sparsely populated. As shown in Table~\ref{tab:gdca_routing_behavior}, GDCA reduces the frequency with which the minority-group fraction falls below $20\%$ from $20.0\%$ to $4.4\%$, while increasing its mean fraction from $29.7\%$ to $37.5\%$. These results indicate that GDCA mitigates routing degeneracy and maintains sufficiently populated groups for more reliable group-specific temperature scaling.

\clearpage

\section{Catastrophic Forgetting and Asymmetric Calibration Behaviors}
\label{sec:forgetting_asymmetric_calibration}
\begin{table}[ht]
\caption{Average first-task performance degradation over 5 runs under the CIL setup.}
\centering
\resizebox{0.8\linewidth}{!}{
\begin{tabular}{lcccccc}
\toprule
Dataset &
\multicolumn{2}{c}{After First Task} &
\multicolumn{2}{c}{After Middle Task} &
\multicolumn{2}{c}{After Final Task} \\
\cmidrule(lr){2-3} \cmidrule(lr){4-5} \cmidrule(lr){6-7}
 & Acc (\%) & Conf (\%) & Acc (\%) & Conf (\%) & Acc (\%) & Conf (\%) \\
\midrule
CIFAR-10
& $ 98.79 $ & $ 96.06 $
& $ 72.69 $ & $ 92.09 $
& $ 20.37 $ & $ 91.38 $ \\

CIFAR-100
& $ 85.62 $ & $ 92.60 $
& $ 42.38 $ & $ 79.14 $
& $ 30.84 $ & $ 75.08 $ \\

Tiny-ImageNet
& $ 70.60 $ & $ 88.47 $
& $ 9.10 $ & $ 68.69 $
& $ 4.70 $ & $ 69.82 $ \\
\bottomrule
\end{tabular}
}
\label{tab:first_task_forgetting}
\end{table}

\begin{figure}[ht]
    \centering
    \begin{subfigure}[t]{0.48\linewidth}
        \centering
        \includegraphics[width=\linewidth]{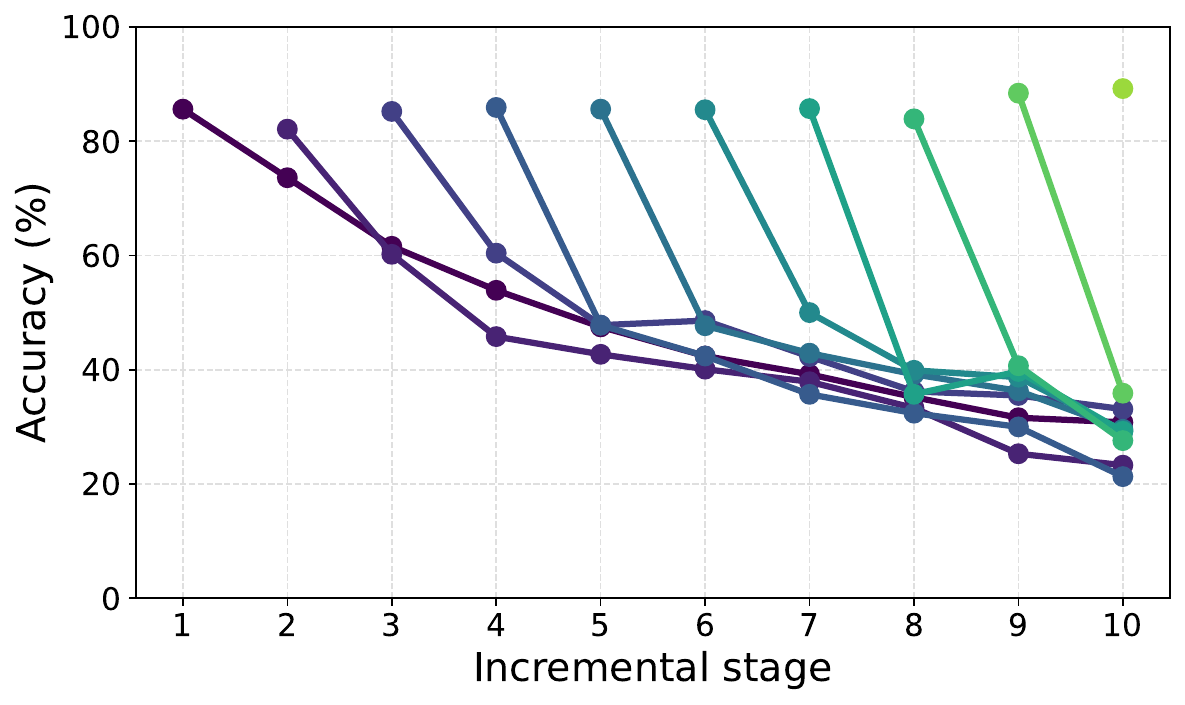}
        \caption{
        Per-task accuracy across incremental stages.
        Accuracy degradation follows different trajectories across tasks,
        indicating asymmetric forgetting dynamics.
        }
        \label{fig:per_task_acc}
    \end{subfigure}
    \hfill
    \begin{subfigure}[t]{0.48\linewidth}
        \centering
        \includegraphics[width=\linewidth]{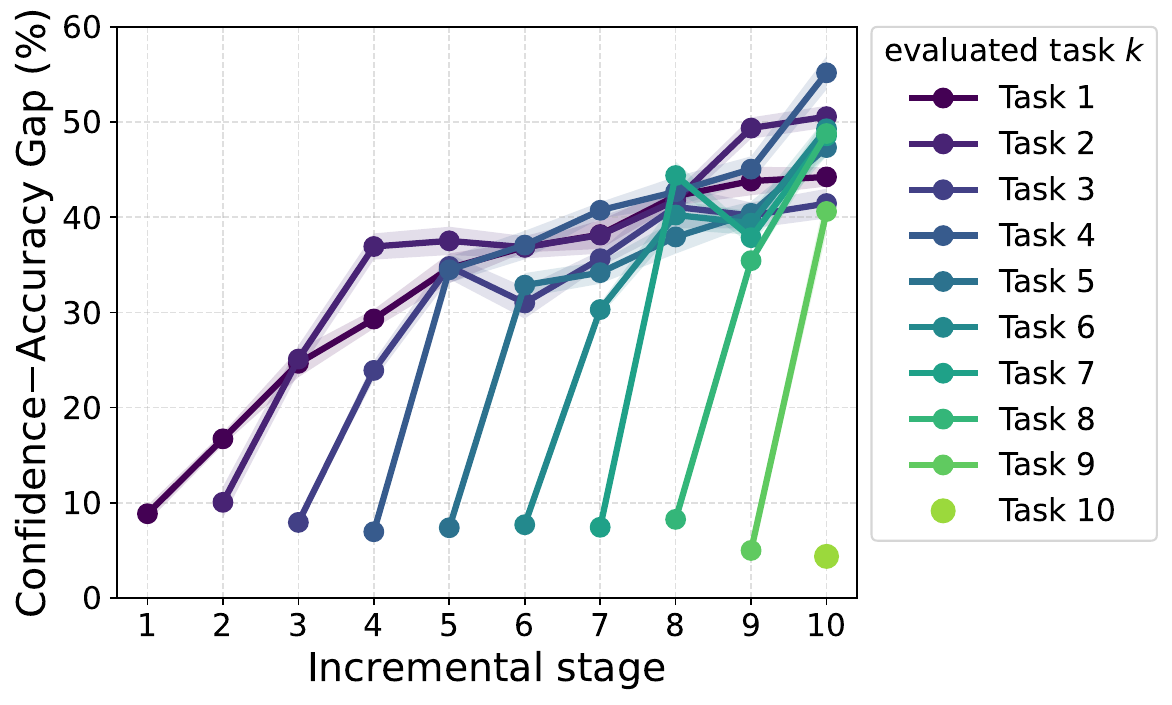}
        \caption{
        Per-task confidence--accuracy gaps.
        The gaps generally widen at different rates across tasks,
        revealing asymmetric calibration behaviors.
        }
        \label{fig:per_task_calibration_gap}
    \end{subfigure}
    \caption{
    Catastrophic forgetting and asymmetric calibration behaviors
    on CIFAR-100. Each curve tracks a task from its introduction
    through subsequent incremental stages.
    }
    \label{fig:forgetting_and_calibration}
\end{figure}

\noindent
Table~\ref{tab:first_task_forgetting} shows that catastrophic forgetting
is accompanied by a growing mismatch between confidence and accuracy.
For the first task of CIFAR-10, accuracy decreases from 98.79\% to
20.37\%, whereas confidence decreases only from 96.06\% to 91.38\%.
Consequently, the confidence--accuracy gap reaches 71.01 percentage
points after the final task.
CIFAR-100 and Tiny-ImageNet exhibit the same pattern:
accuracy declines substantially more than confidence.
These results show how forgetting can contribute to increasing
overconfidence when confidence does not decrease in proportion
to the loss of predictive accuracy.

Figure~\ref{fig:per_task_acc} further shows that the trajectories
of accuracy degradation differ across tasks.
Figure~\ref{fig:per_task_calibration_gap} complements this analysis
by directly tracking the resulting confidence--accuracy gaps
for all tasks on CIFAR-100.
The gaps generally increase after subsequent tasks are introduced,
but the increase is neither uniform nor strictly monotonic:
different tasks exhibit distinct rates of increase and trajectories.
Thus, the mismatch between confidence and accuracy evolves
differently across tasks under continual updates,
revealing asymmetric calibration behaviors.

Taken together, these results connect catastrophic forgetting
with asymmetric calibration: predictive accuracy deteriorates
without a proportional decrease in confidence, and the resulting
gap evolves differently across tasks.
This motivates the use of multiple temperatures in MT-CC
to accommodate heterogeneous calibration needs.

\clearpage
\section{Training and Inference Procedure of MT-CC}

\begin{algorithm}[ht]
\caption{Training Procedure of MT-CC}
\label{alg:training_algorithm}
\centering
\resizebox{0.9\columnwidth}{!}{
\begin{minipage}{\columnwidth}
\footnotesize
\begin{algorithmic}[1]

\Input{
Task sequence $\{\mathcal{D}^1,\dots,\mathcal{D}^B\}$;
calibration sets $\{\mathcal{C}^1,\dots,\mathcal{C}^B\}$;
number of groups $M$
}
\Output{
Grouping network $g_{\phi}$ and group temperatures
$\{T_m^{*}\}_{m=1}^{M}$
}

\State Initialize grouping network $g_{\phi}$
and group temperatures $\{T_m^{0}\}_{m=1}^{M}$

\For{$b = 1$ \textbf{to} $B$}

    \State Split $\mathcal{C}^{b}$ into two disjoint subsets
    $\mathcal{C}^{b} \to
    (\mathcal{S}^{b}_{\mathrm{grp}},
    \mathcal{S}^{b}_{\mathrm{cal}})$

    \State Initialize
    $\{T_m^{b}\}_{m=1}^{M}
    \gets
    \{T_m^{b-1}\}_{m=1}^{M}$

    \State For each $x \in \mathcal{S}^{b}_{\mathrm{grp}}$,
    compute
    $v(x) =
    [\,\delta,
    p_{\mathrm{old}},
    p_{\mathrm{new}},
    \mathbf{p}\,]$

    \State Update $g_{\phi}$ and
    $\{T_m^{b}\}_{m=1}^{M}$
    by minimizing
    $\mathcal{L}_{\mathrm{total}}
    =
    \mathcal{L}_{\mathrm{MT\text{-}CC}}
    +
    \lambda \mathcal{L}_{\mathrm{GDCA}}$

    \State Assign each
    $x \in \mathcal{S}^{b}_{\mathrm{cal}}$
    to
    $m^{*}(x)
    =
    \arg\max_{m} g_{\phi}(x)_m$

    \State For each group $m$,
    refine $T_m^{b}$ on its assigned samples
    in $\mathcal{S}^{b}_{\mathrm{cal}}$
    by minimizing NLL

\EndFor

\State \Return $g_{\phi}$ and
$\{T_m^{B}\}_{m=1}^{M}$

\end{algorithmic}
\end{minipage}
}
\end{algorithm}

\textbf{Training.}
Algorithm~\ref{alg:training_algorithm} summarizes MT-CC training.
For each task, calibration data are split into grouping and calibration subsets.
We jointly optimize the grouping network and group temperatures using $\mathcal{L}_{\text{total}}$, then freeze $g_{\phi}$ and refine each group temperature with NLL. The resulting temperatures initialize the next task.

\begin{algorithm}[ht]
\caption{Inference Procedure of MT-CC}
\label{alg:inference_algorithm}
\centering
\resizebox{0.9\columnwidth}{!}{
\begin{minipage}{\columnwidth}
\footnotesize
\begin{algorithmic}[1]
\Input{Grouping network $g_{\phi}$, optimized group temperatures $\{T_m^{*}\}_{m=1}^{M}$, test sample $x$ and global temperature $T_{\text{glob}}^{*}$}
\Output{Calibrated probability vector $p_{\mathrm{cal}}(x)$}

\If{$b = 1$}
    \State $p_{\mathrm{cal}}(x) \leftarrow \mathrm{softmax}\!\left(\frac{f_{\theta}(x)}{T_{\text{glob}}^{*}}\right)$
\Else
    \State $g^{*}(x) \leftarrow \arg\max_{m} g_{\phi}(x)_m$
    \State $p_{\mathrm{cal}}(x) \leftarrow \mathrm{softmax}\!\left(\frac{f_{\theta}(x)}{T_{g^{*}(x)}^{*}}\right)$
\EndIf
\end{algorithmic}
\end{minipage}
}
\end{algorithm}

\noindent\textbf{Inference.}
\noindent Algorithm~\ref{alg:inference_algorithm} summarizes the inference procedure of MT-CC.
For the first task, the logits $f_{\theta}(x)$ are calibrated using a single optimized global temperature $T_{\text{glob}}^{*}$ through standard temperature scaling.
For subsequent tasks, the grouping network $g_{\phi}$ assigns each test sample to the group with the highest predicted probability.
The calibrated confidence is then obtained by applying the corresponding optimized group-specific temperature to the logits before the softmax operation.

\clearpage
\section{Reliability Diagram}

\begin{figure}[ht]
    \centering

    % ---------- 1행 : T-CIL ----------
    \begin{subfigure}[b]{0.25\linewidth}
        \centering
        \includegraphics[width=\linewidth]{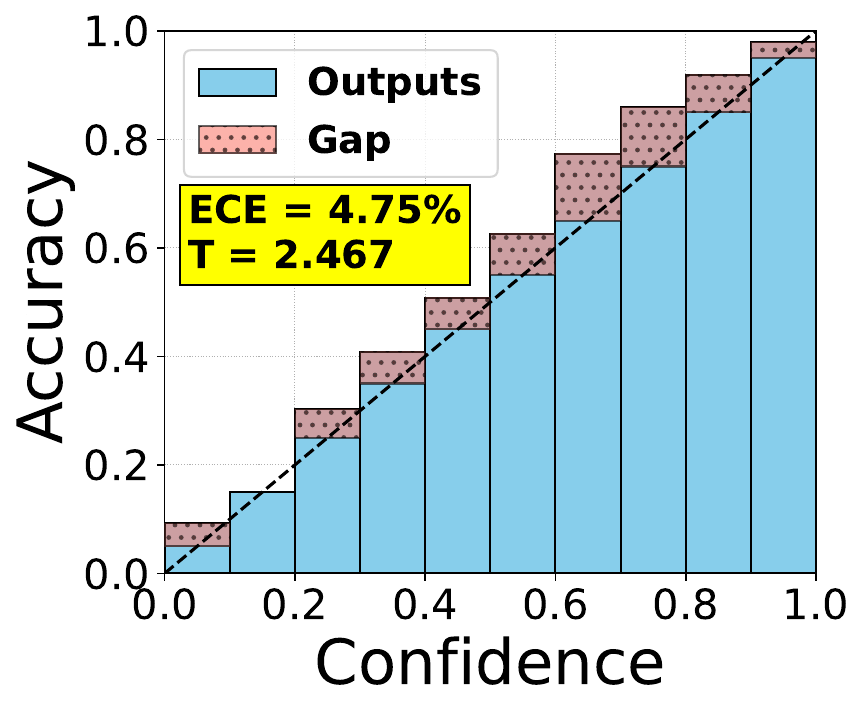}
        \caption{Single $T$ (group 1)}
        \label{fig:tcil_g1}
    \end{subfigure}
    \begin{subfigure}[b]{0.25\linewidth}
        \centering
        \includegraphics[width=\linewidth]{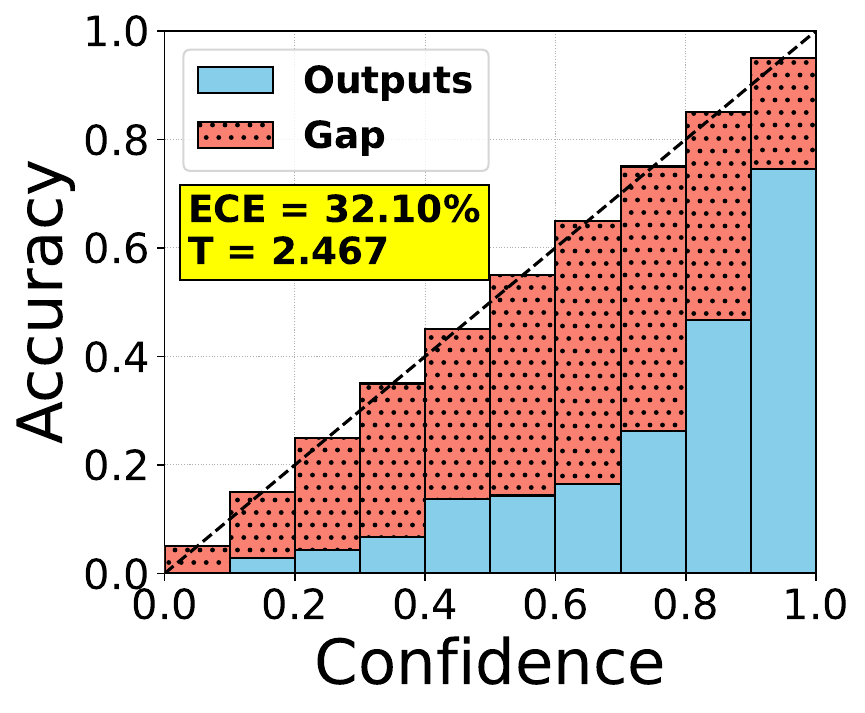}
        \caption{Single $T$ (group 2)}
        \label{fig:tcil_g2}
    \end{subfigure}
    \begin{subfigure}[b]{0.25\linewidth}
        \centering
        \includegraphics[width=\linewidth]{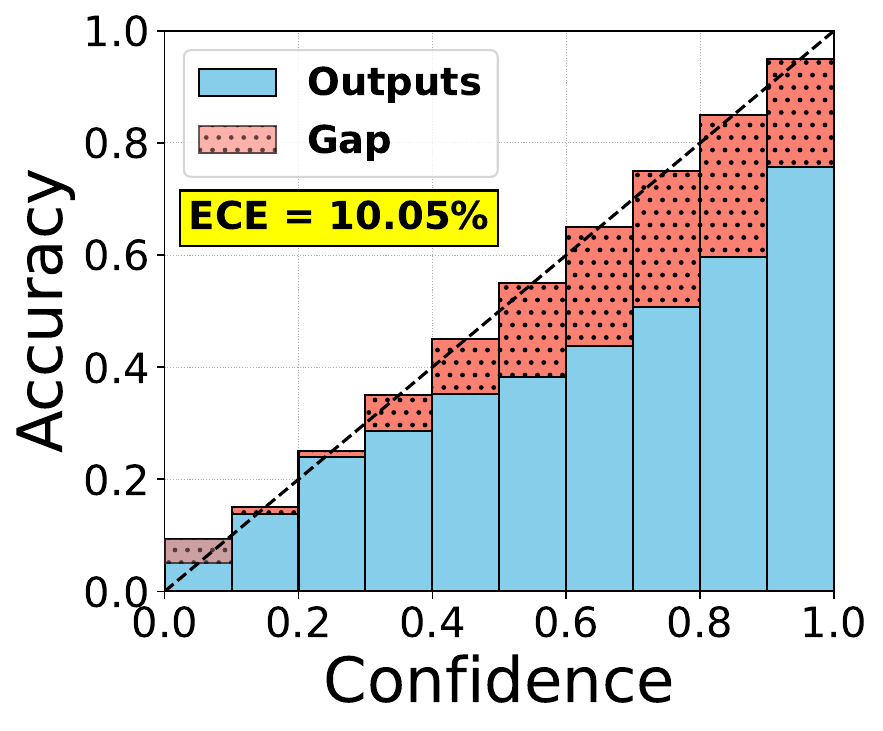}
        \caption{Single $T$ (overall)}
        \label{fig:tcil_all}
    \end{subfigure}

    \par\medskip

    % ---------- 2행 : MT-CC ----------
    \begin{subfigure}[b]{0.25\linewidth}
        \centering
        \includegraphics[width=\linewidth]{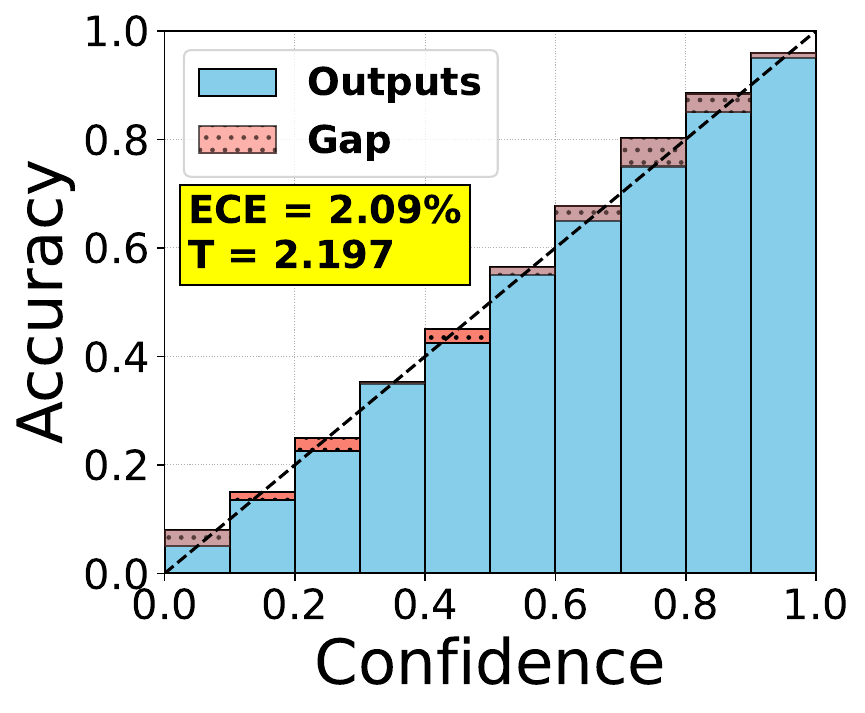}
        \caption{MT-CC (group 1)}
        \label{fig:mtcc_g1}
    \end{subfigure}
    \begin{subfigure}[b]{0.25\linewidth}
        \centering
        \includegraphics[width=\linewidth]{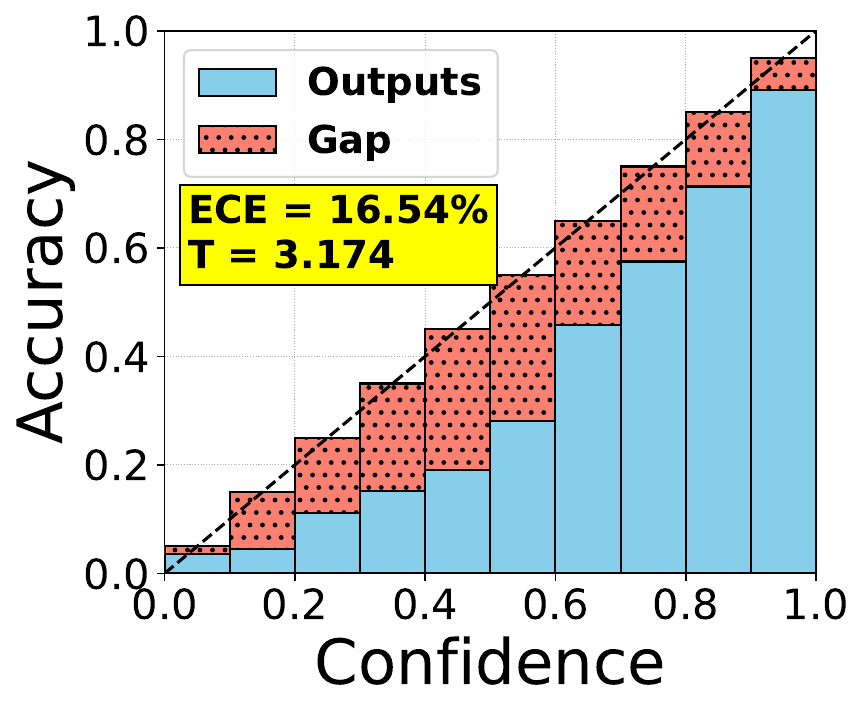}
        \caption{MT-CC (group 2)}
        \label{fig:mtcc_g2}
    \end{subfigure}
    \begin{subfigure}[b]{0.25\linewidth}
        \centering
        \includegraphics[width=\linewidth]{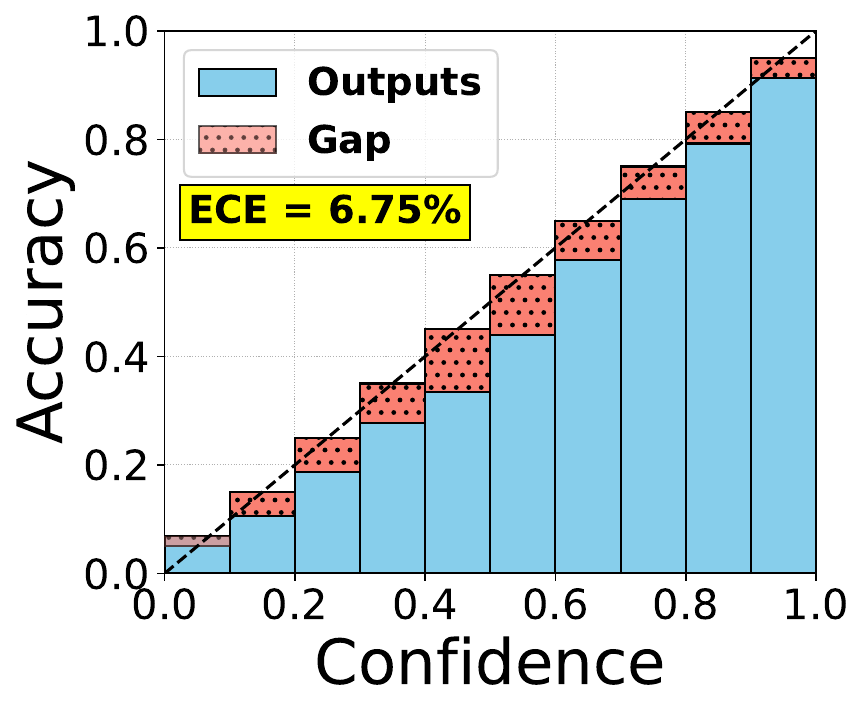}
        \caption{MT-CC (overall)}
        \label{fig:mtcc_all}
    \end{subfigure}
    \caption{
        \textbf{(a), (b), (c):} Reliability diagrams of T-CIL~\citep{hwang2025t} using a single temperature $T$.
        \textbf{(d), (e), (f):} Reliability diagrams of the proposed MT-CC with optimized grouping.
        Results are shown after training on the 10th task of CIFAR-100, comparing group 1, group 2, and all classes as determined by the proposed grouping function.
    }
    \label{fig:reliability_mtcc_tcil_task10}
\end{figure}

\subsection{Group-wise Reliability Diagrams at the Final Task}

Figure~\ref{fig:reliability_mtcc_tcil_task10} compares the reliability diagrams of T-CIL and MT-CC after training on the 10th (final) task of CIFAR-100, separately for group 1, group 2, and the overall distribution determined by the proposed grouping network.
With a single global temperature, T-CIL exhibits clearly distinct miscalibration patterns between the two groups: one group remains overconfident while the other becomes underconfident, so a single $T$ cannot satisfy both simultaneously, leaving residual gaps in the overall reliability diagram.
In contrast, MT-CC assigns a group-specific temperature to each subset of samples, substantially reducing the gap between confidence and accuracy within both groups and yielding a tighter alignment with the diagonal in the overall reliability diagram.
This visually confirms our central claim that addressing the asymmetric calibration behaviors of CIL with multiple temperatures leads to more effective calibration than the single-temperature strategy.

\subsection{Reliability Diagrams across Sequential Tasks}

Figure~\ref{fig:cifar100_reliability} presents the reliability diagrams for CIFAR-100 from Task 2 to Task 10, comparing T-CIL and MT-CC across Group 1, Group 2, and the overall distribution.
The diagrams illustrate how closely the model’s predicted confidence aligns with the empirical accuracy, where curves closer to the diagonal indicate better calibration.
Across most tasks, MT-CC exhibits consistently stronger overall calibration than T-CIL, confirming its advantage.

\newcommand{\RelRow}[1]{%
    % 내부에서 (#1-1) 을 계산한 숫자를 만든다
    \edef\FileNum{\the\numexpr #1 - 1\relax}%

\begin{tikzpicture}
\node (img) {
    \begin{tabular}{c@{}c@{}c@{}c@{}c@{}c}
        % --- T-CIL ---
        \includegraphics[width=0.154\textwidth]{Figure/T_CIL_group0_reliability_task\FileNum.pdf} &
        \includegraphics[width=0.154\textwidth]{Figure/T_CIL_group1_reliability_task\FileNum.pdf} &
        \includegraphics[width=0.154\textwidth]{Figure/T_CIL_all_class_reliability_task\FileNum.pdf} &
        % --- MT-CC ---
        \includegraphics[width=0.154\textwidth]{Figure/MT_CC_group0_reliability_task\FileNum.pdf} &
        \includegraphics[width=0.154\textwidth]{Figure/MT_CC_group1_reliability_task\FileNum.pdf} &
        \includegraphics[width=0.154\textwidth]{Figure/MT_CC_all_class_reliability_task\FileNum.pdf}
    \end{tabular}
};
\draw[dashed, thick] ([yshift=-2mm]img.north) -- ([yshift=2mm]img.south);
\node[rotate=90, anchor=center]
      at ([xshift=0cm]img.west)
      {\textbf{Task #1}};
\end{tikzpicture}%
}
% Header that mirrors \RelRow's tikz structure so it shares the same total
% width as the image rows below (otherwise the rotated "Task X" label on the
% left of \RelRow shifts the image grid right and headers go out of alignment).
\newcommand{\RelHeader}{%
\begin{tikzpicture}
\node (hdr) {%
  \begin{tabular}{c@{}c@{}c@{}c@{}c@{}c}
    \multicolumn{3}{c}{\textbf{T-CIL}} & \multicolumn{3}{c}{\textbf{MT-CC}} \\[1pt]
    \makebox[0.154\textwidth]{Group 1} &
    \makebox[0.154\textwidth]{Group 2} &
    \makebox[0.154\textwidth]{Overall} &
    \makebox[0.154\textwidth]{Group 1} &
    \makebox[0.154\textwidth]{Group 2} &
    \makebox[0.154\textwidth]{Overall}
  \end{tabular}%
};
\draw[dashed, thick] ([yshift=-1mm]hdr.north) -- ([yshift=1mm]hdr.south);
% Invisible rotated "Task X" placeholder that takes up the same width as the
% rotated label in \RelRow, so the image grid and header line up exactly.
\node[rotate=90, anchor=center]
      at ([xshift=0cm]hdr.west)
      {\phantom{\textbf{Task 10}}};
\end{tikzpicture}%
}
\begin{figure}[t]
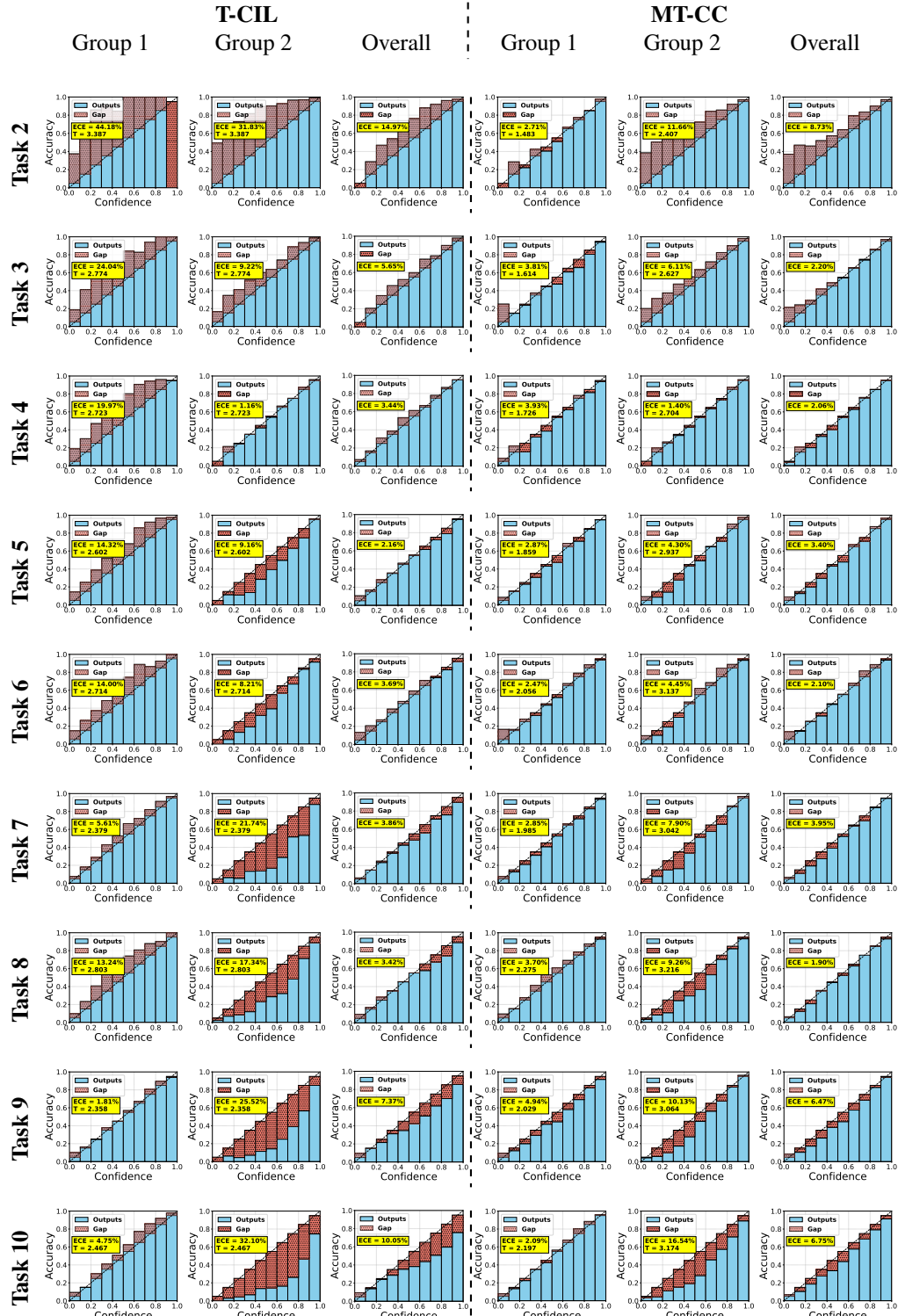

\centering
\begin{tabular}{c}
  \RelHeader \\
  \vspace{-1.5mm}
  \RelRow{2} \\
  \vspace{-1.5mm}
  \RelRow{3} \\
  \vspace{-1.5mm}
  \RelRow{4} \\
  \vspace{-1.5mm}
  \RelRow{5} \\
  \vspace{-1.5mm}
  \RelRow{6} \\
  \vspace{-1.5mm}
  \RelRow{7} \\
  \vspace{-1.5mm}
  \RelRow{8} \\
  \vspace{-1.5mm}
  \RelRow{9} \\
  \vspace{-1.5mm}
  \RelRow{10}
\end{tabular}
\vspace{-5mm}
\caption{Reliability diagrams for CIFAR-100 comparing T-CIL and MT-CC across Tasks 2--10.
Overall calibration performance shows consistently strong reliability across most tasks.}
\label{fig:cifar100_reliability}
\end{figure}
\clearpage

\end{document}